\documentclass[letterpaper, 10 pt, journal]{IEEEtran}  

\usepackage[utf8]{inputenc} 
\usepackage[T1]{fontenc}    

\usepackage{url}            
\usepackage{booktabs}       
\usepackage{amsfonts}       
\usepackage{nicefrac}       
\usepackage{microtype}      
\usepackage{moreverb,url}

\usepackage[hidelinks]{hyperref}

\usepackage{amsmath} 
\usepackage{amssymb} 
\usepackage{graphicx} 
\graphicspath{{figures/}{figures/}}
\usepackage{subfig}
\usepackage{caption}
\usepackage[calcwidth = \columnwidth]{caption}
\usepackage{booktabs,threeparttable} 
\usepackage{setspace} 
\usepackage{mathptmx}
\usepackage{textcomp}
\usepackage{float} 
\usepackage{rotating}
\usepackage{wrapfig}
\usepackage{color}
\usepackage{algorithm}
\usepackage[noend]{algpseudocode}
\usepackage{multirow}
\usepackage{eqnarray}
\usepackage{hhline}

\usepackage[symbol]{footmisc}

\usepackage{epsfig} 
\usepackage{mathtools}
\usepackage{tabularx}
\usepackage{array}

\begin{document}

\title{Human-robot co-manipulation of extended
objects: Data-driven models and control from analysis of
human-human dyads}


\author{Erich Mielke$^1$, Eric Townsend$^1$, David Wingate$^2$, and Marc D. Killpack$^1$ 
\thanks{Authors are affiliated with the Mechanical Engineering (1) and Computer Science (2) departments at Brigham Young University, Provo, UT, 84602 USA}}

\markboth{Pre-print Paper, Submitted November 2018}%
{}

\maketitle

\begin{abstract}
Human teams are able to easily perform collaborative manipulation tasks. However, for a robot and human to simultaneously manipulate an extended object is a difficult task using existing methods from the literature. Our approach in this paper is to use data from human-human dyad experiments to determine motion intent which we use for a physical human-robot co-manipulation task. We first present and analyze data from human-human dyads performing co-manipulation tasks.  We show that our human-human dyad data has interesting trends including that interaction forces are non-negligible compared to the force required to accelerate an object and that the beginning of a lateral movement is characterized by distinct torque triggers from the leader of the dyad. We also examine different metrics to quantify performance of different dyads. We also develop a deep neural network based on motion data from human-human trials to predict human intent based on past motion.  We then show how force and motion data can be used as a basis for robot control in a human-robot dyad. Finally, we compare the performance of two controllers for human-robot co-manipulation to human-human dyad performance. 
\end{abstract}


\section{Introduction}
In the future, robots will work alongside humans in many applications including logistics, health-care, agriculture, disaster response, and others. The advantage of human-robot collaboration in these areas is that humans provide intelligence and dexterity while robots may provide strength, stability, and even redundancy \cite{Kazerooni1990}. Physical Human-Robot Interaction (pHRI) for collaborative manipulation (or co-manipulation) is an area of robotics that can especially benefit from the combined strengths of a human-robot team: strength and execution from the robot and intelligence and planning from the human. This is particularly true of co-manipulation tasks where a human and a robot physically manipulate the same object simultaneously. Co-manipulation can include complex translational and rotational tasks, such as moving a table, couch, or other extended, rigid objects. These objects may be heavy or unwieldy, which could necessitate two or more people to carry them. A robot capable of replacing a human in these teams would help in situations like search and rescue where current high-payload robots are too heavy and dangerous to relocate and operate. Robots that can physically interact with a human could help lift and remove rubble from disaster areas or take a victim on a stretcher to safety. These robots would allow fewer people to complete the same amount of work, or for more teams to operate and reach more people in need of help. Other applications include using robots to help load and unload moving vans, using robots to help move objects around warehouses, and any other co-manipulation applications where human-human teams are currently needed. 

In these situations, robots will need to work safely and intuitively, in order to be an asset when interacting with people. Specifically, they will need to be able to predict and respond to human intent in an effective manner. However, an important characteristic of these situations is uncertainty in the task. Often a task is poorly defined for one or both partners of a dyad, and a controller needs to be able to adapt to disturbances and trajectory changes. Uncertainty and ambiguity can especially exist when tasks include manipulating an extended object that may need to be translated, rotated, or both. When an extended object is included in co-manipulation tasks, forces applied in a lateral direction could indicate either intent to translate laterally, or intent to rotate the object in the plane, which will be referred to as the rotation-translation problem. In order to be effective, a pHRI controller for co-manipulation of extended objects must be able to distinguish between an intent to rotate and translate. One method of assessing human intent is to study human-human interactions (HHI). By studying HHI data, we can define patterns or characteristics that will help to create a safe and intuitive co-manipulation controller. Therefore, this paper proposes a method for predicting human intent in a co-manipulation task based on HHI. For clarity, we have designated human intent as the intent to move an object in a particular direction with a particular velocity.

\begin{figure}[t]
  \centering
  \includegraphics[width=.9\linewidth]{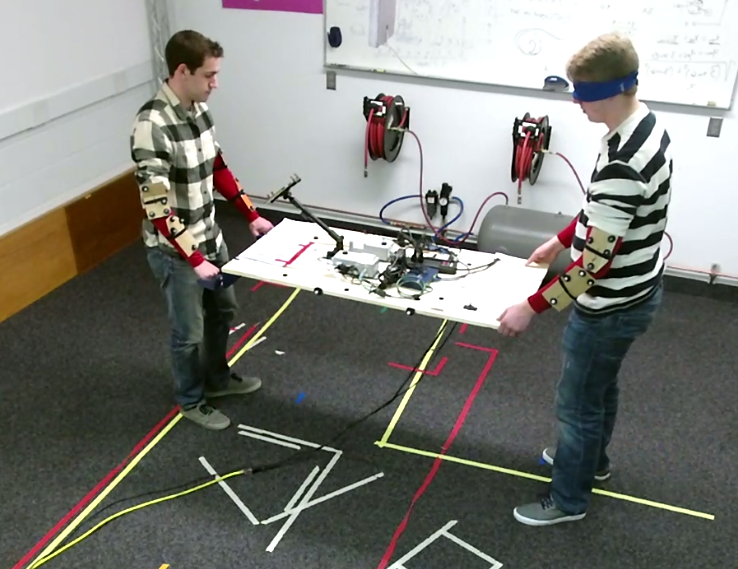}
  \caption{A leader and a blindfolded follower performing a table carrying task.}
  \label{fig:dyad}
\end{figure}

In order to understand human-human co-manipulation of rigid, extended objects, we ran an exploratory study with 21 human dyads. Each dyad moved a long board representing a table as we measured their motion and forces on the board as in Fig. \ref{fig:dyad}. Although other studies have been completed analyzing human movement, many of these are done in haptic simulations or with limited degrees of freedom in order to isolate specific behaviors. These studies have given significant insight on things like minimum jerk motion, negotiation of roles, and task-specific movements. Due to the nature of these past studies, which have mostly examined a limited number of degrees of freedom, there are limitations to how those results can be extrapolated for general purpose six dimensional co-manipulation tasks. The HHI study described in detail in this paper (and as first described in \cite{Mielke2017}) is necessary to study how motion is affected without these limitations, and allows us to validate what has been learned in other studies and allows for further insight and direction for human-robot co-manipulation controller development.

There are many sources of information that a robot could use to
predict human intent, including motion, force, partner posture, and verbal communication among others. In our study, we chose to focus on motion and force, since we believed that these variables were the most
fundamental and easiest for a robot to interpret in order to predict
what the person intends to do. This does not mean that other information sources could not be used to improve upon our results, but rather that this data is sufficient to characterize human intent in co-manipulation tasks. Other studies have confirmed that haptic-channel communication is sufficient to indicate motion intent \cite{Sawers2017}. However, while the past work on co-manipulation outlined in Section \ref{related_Work} shows that collaboration through force is applicable to some tasks, it is not clear that previously developed algorithms and intent-estimators will work in less-defined scenarios. This is especially true since co-manipulation is likely to include whole-body motion, bi-manual manipulation by the participants, six degree of freedom motion of the object, rather than planar arm movements only. By basing our human-intent model on data from human-human dyads, we are increasing the likelihood that our controller will be intuitive for human users. The initial goal of our co-manipulation controllers is to know how the robot should move, based on sensory inputs, in order to manipulate the object being carried in the manner desired by the human partner. 

The specific contributions of this paper include the following:
\begin{enumerate}
	\item Unique co-manipulation data from human-human dyads moving a rigid table (see Section \ref{human_study})
	\item Observations on planar movements from human-human co-manipulation study (see Section \ref{obsv}), which include the following:
	\begin{itemize}
		\item Interaction forces are not minimized
		\item Trajectories resemble minimum-jerk trajectories
		\item Correlation between dyads that minimized completion time and dyads that minimized deviation from a minimum-jerk trajectory
		\item Lateral movements are triggered by a specific torque sequence
		\item Planar rotation movements can be distinguished from lateral movements by differing torque sequence triggers		
	\end{itemize}
    \item Development of a neural network to predict human intent based on past motion of the human dyad (see Section \ref{sec:neural-network}).
    \item Application of neural network and trigger based predictions to a human-robot dyad, comparing performance of human-robot dyads with human-human dyads (see Section \ref{sec:results-and-discussion}).
\end{enumerate}

The organization of the remainder of this paper is outlined next. Section \ref{related_Work} describes related work on physical human-robot interaction and intent modeling. Next, the human-human dyad experiment is explained in Section \ref{human_study}, including a description of the equipment, the tasks performed, and the participants. Section \ref{obsv} then explores the main observations of the study. In Section \ref{evic} we discuss the formulation and preliminary testing of our Extended Variable-Impedance Controller for human-robot co-manipulation. We then describe the structure, training, and validation of the neural network, as well as the formulation of a neural-network based controller in Section \ref{sec:neural-network}. In Section \ref{sec:pilot} we describe a physical human-robot co-manipulation experimental study with both human-robot controllers. Finally we discuss the results of the human-robot study in Section \ref{sec:results-and-discussion} with conclusions in Section \ref{sec:conclusion}.  

\section{Related Work}
\label{related_Work}
Researchers have been studying aspects of pHRI co-manipulation for many years. We have grouped the efforts of these researchers into a few different categories: studies about co-manipulation or human behaviors, force-based and motion-based co-manipulation methods, determining performance of human-robot dyads through metrics, and human intent estimation. 

\subsection{Co-Manipulation and Human Behavior Studies}
One of the most widely used studies that explores human-arm reaching movement was performed by Flash and Hogan \cite{Flash1985}. It illustrates the tendency of upper-arm reaching movements to resemble minimum-jerk trajectories. Another fundamental study was performed by Rahman et al. \cite{Rahman2000} where they performed a 1 DOF translation co-manipulation experiment between two human users. They showed showed that the stiffness and damping parameters changed over the course of the task, which is known as variable impedance. They also showed that for 1 DOF translation tasks, the achieved trajectory corresponded well with the minimum-jerk trajectory. 

There were also a number of studies investigating how humans cooperate through forces and haptic channels. Reed et al. \cite{Reed2007} performed an experiment testing their hypothesis that humans can cooperate by specializing their forces. They found that human-human dyads were able to perform simple tasks significantly faster than they did when working alone. Ganesh et al.'s study \cite{Ganesh2014} also showed that a physical connection between two individuals improved the dyad's performance. In a different study by Wel et al. \cite{Wel2011}, this result was confirmed as subjects performing a 1 DOF task benefited from the added haptic channel communication. However, when Reed et al. included a robot, this advantage disappeared. 

Other studies show that not only does a haptic channel improve performance of a human-human dyad under normal conditions, but that a haptic channel can be used as the only source of information exchange between partners. Sawers et al. \cite{Sawers2017} performed an experiment where participants performed a series of dance steps with a partner. Another study by Mojtahedi et al. \cite{Mojtahedi2017} also showed that interaction forces may communicate movement goals between human-human dyads in cooperative physical interactions. 

One of the only studies performed with a human-human dyad carrying an extended object was done by Bussy et al. \cite{Bussy2012a}. In this experiment, they had dyads move a beam in 1 DOF, forward and backward and used object velocity to trigger state transitions in a state machine model. 

\subsection{Control Methods for Co-Manipulation}

\subsubsection{Force-Based Co-Manipulation Methods}
A large portion of the approach for intent estimation and robot control presented in this paper is based on different methods presented for human-robot co-manipulation over the last 20 years. However, it is not clear how many of these past control methods apply to large-scale, extended objects, or to movements requiring more than 1 or 2 DOF. One of the first controllers for cooperative manipulation of an object by robots and humans was an impedance controller developed by Ikeura et al. \cite{Ikeura2002,Ikeura}. They also developed strategies for situations that required using direction of force and change in magnitude of force. This type of control technique is known as variable-impedance control \cite{Ikeura1965,Dimeas2015}. The defining characteristic of this method is measuring Cartesian-coordinate forces at the end effector to determine motion intent in certain Cartesian directions. Tsumugiwa et al. \cite{TSUMUGIWA2002} showed that varying the impedance allows for increased performance of human-robot interaction in calligraphy. This variable impedance approach was also very successful in predicting Cartesian movements, as was shown in other studies as well, \cite{Duchaine2007,Ficuciello2015}. However, it does not generalize to include rotational movements. It also is heavily dependent on human force input, meaning the robot does not proactively contribute to moving the object being manipulated, and the human partner must exert more force than may be required in a human-human dyad. 
HERE
The initial work in variable impedance control (VIC), however, provided a basis for using haptic information in future pHRI controllers. One such controller was implemented by Ranatuga et al. \cite{Ranatunga2016}. They were able to perform 1 DOF point-to-point motion tasks without previous knowledge of the trajectory, which is necessary for situations such as search and rescue. However, the work assumed direct contact between human and robot, i.e. no extended object co-manipulation, and was limited in DOF. In fact, there is an inherent problem with VIC, and other methods, such as Leica et al.'s method for moving extended objects \cite{Leica2013}, that limits how many DOF are viable. This is known as the translation versus rotation (TvR) problem. In a simple planar task, the leader has the option of moving the extended object by either translating forward/backward, translating laterally, or rotating the board. The problem arises when the leader wishes to move laterally, and so applies a force in that direction. The follower, who lies some distance away from the applied force, perceives the force as a torque, and begins to rotate the board. This shows that there is information missing in VIC to deal with the TvR problem.

Two approaches to solve this problem were made by Karayiannidis et al. and Nguyen \cite{Karayiannidis2014,Nguyen2016}. Karayiannidis et al. used the direction and magnitude of the applied force to an extended object to create a state machine which switches between translation and rotation modes. The state machine, however, fails to transition between states correctly when moving at different speeds than described in their experiment. Nguyen improved upon this by using Hidden Markov Models and showed that it is possible to predict human behavior in co-manipulation tasks. The algorithm allowed for different speeds of rotation and translation, but ultimately performed worse than Karayiannidis et al.'s method. Neither compared their controller performance to any of the metrics established by other researchers. 

Other work has been done by Peternel et al. \cite{Peternel2017} where they incorporated EMG sensor feedback with the control law to provide more information about the stiffness the human was applying in a 1 DOF sawing task. Additionally, Peternel et al., in a different work \cite{Peternel2017a}, showed how robots can adapt to human fatigue in pHRI. 

One of the only attempts at bi-manual, planar human-robot co-manipulation was developed by Bussy et al. \cite{Bussy2012}. Their method relied on force inputs to a trajectory-based control law, where the trajectories are then decomposed into a finite state machine to determine the desired velocities. This research was successful in at least anterior translation coupled with planar rotation, and theoretically generalizes to include lateral translation. However, they do not mention attempts to move in lateral translation, and a video of the controller shows only anterior translation with planar rotation. It is therefore unclear how they deal with the TvR problem

\subsubsection{Motion-Based Co-Manipulation Methods}
In addition to force-based methods, many insights into human-robot interaction have been gained from studying motion-based intent. One of the common methods of motion-based co-manipulation is using a minimum-jerk basis. Corteville et al. \cite{Corteville2007}, did so for a 1 DOF point-to-point experiment. Also, Maeda et al. used minimum-jerk trajectories to predict human intent for proactive robot behavior \cite{Maeda2001}. This strategy reduced the amount of effort a human partner needed to exert in co-manipulation tasks, which is one of the problems with variable impedance control. 

However, there are restrictions on using minimum-jerk as a controller basis. The trajectory start and end points must be known beforehand, minimum-jerk based methods are not robust to disturbances, and they do not take into consideration the haptic channels, which have been shown to increase performance. Additionally, Thobbi et al. \cite{Thobbi2011} showed that there are some human movements that are not minimum-jerk movements. They approached solving the TvR problem by learning a model that allowed the robot to account for a larger variation in trajectory. However, they do not consider higher DOF, nor do they incorporate haptic inputs. Miossec and Kheddar \cite{Miossec2008} also explored non-minimum jerk-based trajectories. They continued the work done by Bussy et al. \cite{Bussy2012a}, where dyad motions are longer and include walking and not just arm movement. Their experiment indicated that the interaction forces are too difficult to use as a basis for control, and instead they use velocity thresholds \cite{Bussy2012}. However, because their experiment only looked at 1 DOF, it is not clear how their method would extend to higher DOF. 

Ge et al. \cite{Ge2011} showed that machine learning can be a useful tool in pHRI. Their research used supervised learning to predict the motion of the human limb. While their work, along with that shown by Thobbi et al. \cite{Thobbi2011}, shows that human performance can be learned and applied to pHRI controllers, they did not account for co-manipulation of an extended object. Additionally, they indicated that the interaction force between a human-robot dyad should be minimized, which may not be an ideal objective for some co-manipulation tasks (see Section \ref{obsv}). Another use of machine learning was demonstrated by Berger et al. \cite{Berger2015} where they used accelerometer and pressure sensor information to learn a statistical model to guide the robot's behavior. However, they did not explore the TvR problem, and it is not clear how well this method performed in comparison to human-human dyads. More recently, Lanini et al. \cite{lanini2018human} used a multi-class classifier to determine if a robot should start/stop-walking, accelerate, or decelerate for a seemingly one DoF task with a single arm. 

A different approach was taken by Medina et al., who demonstrated an anticipatory model that took into account expected human trajectory as well as human force variability \cite{Medina2015}. They showed that incorporating uncertainty led to a higher perceived helpfulness in the robot partner of a human-robot dyad. However, the cost function used in this research incorporated a term for human force minimization, which may not necessarily be a goal in co-manipulation (see Section \ref{obsv}). It also was implemented in 2 DOF with a single-arm virtual haptic interface, so it is unclear whether it would extend to bi-manual co-manipulation with an extended object in higher DOF. 

\subsection{Performance Metrics}
An issue in co-manipulation studies and methods is determining what constitutes a successful dyad. One dyad might take longer than the other, or a dyad might also have more variability in motion than another dyad. Therefore, there needs to be performance metrics that allow for comparison between dyads. 

Haptic information has been shown to be a viable communication method, and some researchers have suggested this information is used by dyads to minimize certain criteria. Groten \cite{Groten2011} described a number of these metrics, including minimizing interaction forces and root-mean-square error, and maximizing time on target. A reference trajectory that is commonly used, such as in Corteville et al. \cite{Corteville2007} and other previously mentioned studies, is the minimum-jerk trajectory. However, there are also tasks that do not fit well with the minimum-jerk trajectories \cite{Miossec2008,Thobbi2011}. Therefore, some alternative trajectories may need to be used if using a root-mean-square error on trajectory. 

Ivaldi et al. \cite{Ivaldi2012} also described a few other metrics, such as minimizing jerk, torque change, geodesic trajectories, energy, and effort. These are all fairly well explained by their titles, and the objective of minimizing these metrics is to achieve human-like behavior. More metrics not mentioned by Ivaldi et al., but commonly used in other works are minimizing task completion time \cite{Duchaine2007,Miossec2008} and position error in trajectory following tasks such as tracing a path through a maze \cite{Thobbi2011,Ikeura1965}. In reviewing the literature, it is not clear which of these metrics is most important. For example, completion time can capture how quickly a dyad is able to complete a task, but this may not be the most important measure of the dyad. Perhaps their task involves moving as close to the desired trajectory as possible, and moving faster causes more errors. In this paper we propose a few additional metrics and we use this list of metrics as a basis for characterizing performance of co-manipulation methods. Determining what behavior dyads display is essential not only for comparing one dyad to another, but also for comparing one co-manipulation controller to another.

\subsection{Human Intent Estimation}
One of the main hurdles remaining in human-robot co-manipulation is effective human intent estimation. Many papers have suggested that haptic channels are an appropriate method of communication for human intent \cite{Basdogan2001,Noohi2016,Reed2007,Groten2013}. This makes sense, as we have seen that human teams can move objects by interacting only through forces applied to the objects, rather than by communicating verbally or otherwise \cite{Sawers2017,Mojtahedi2017}. Many studies have been done to conclude that robots can be controlled by human force input in this manner, but these studies often involve the human acting directly on the robot, and not through any extended object \cite{Corteville2007,Ikeura1997,Ikeura,TSUMUGIWA2002}.

Some past research involved shared virtual-environment loads \cite{Madan2015,Lawitzky2012}, and others involved upper-arm movements of individuals and dyads \cite{Wel2011,Reed2007,lanini2018human}. These experiments clarified many aspects of pHRI, including verifying that haptic information aids in co-manipulation tasks, noting some interaction patterns, and combining planning and learning to complete goal-oriented tasks. However, only exploring virtual environments or upper-arm movements may oversimplify the problem of general six degree of freedom co-manipulation. 

Another method of intent estimation that has been used is programming by demonstration, as in Rozo et al. \cite{Rozo2016}. Here, intent is compressed into a section of possible motions the human-robot dyad could take. The disadvantage is that it is not robust to disturbances, or trajectories that have not been previously modeled.

Our definition of intent is appropriate for co-manipulation of extended objects because it allows us to capture intent for motion that does not have definite start of end points (as observed by the robot), or motion that involves unforeseen obstacles.

\subsection{Related Work Summary}

The works cited here describe most of the current research being performed in pHRI co-manipulation. As has been shown, there are very few studies that look at co-manipulation of extended objects, and even fewer that look at high DOF bi-manual  co-manipulation. Approaches for control methods are varied between force-based and motion-based, but almost all are limited in applicability due to low DOF, or lack of generality (requiring previous knowlege about a desired trajectory). We also have not seen a working bi-manual co-manipulation controller for a human-robot dyad, with at least 3 DOF that can be used in undefined situations or respond to disturbances, in any of the related literature.

\section{Human Dyad Experiment}\label{human_study}

\begin{figure}[t]
  \centering
  \includegraphics[width=1.0\linewidth]{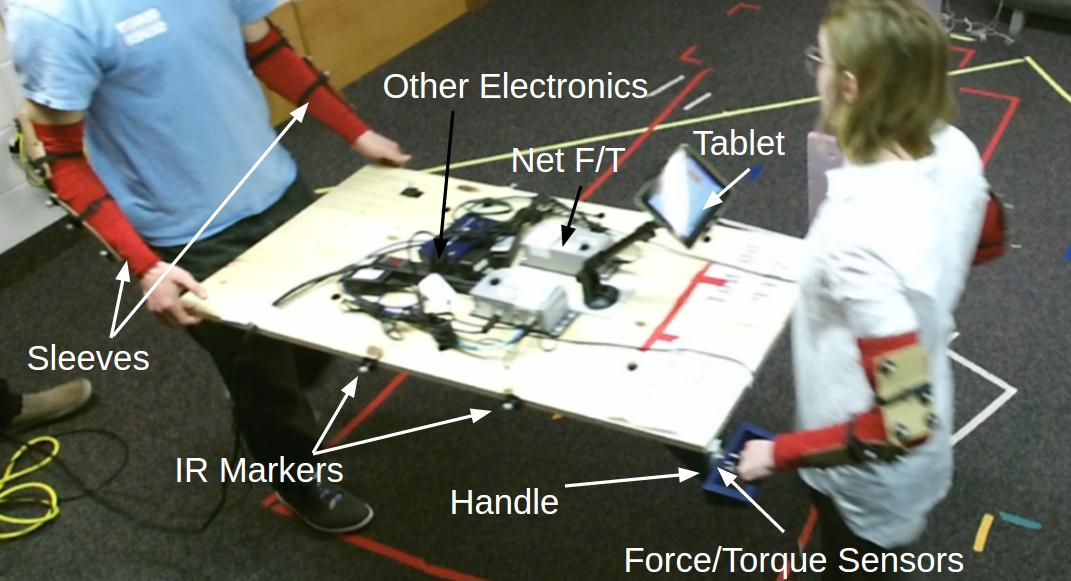}
  \caption{Setup for table and during trials}
  \label{fig:TableConfig}
\end{figure}

\begin{sloppypar}
Our human-human co-manipulation experiment involved tasks ranging from 1 DOF up to 6 DOF. The purpose of this study was two-fold: first, to provide a baseline for how humans perform a general collaboration task with extended objects, and second, to provide useful haptic information to use for creation of a human intent estimator. Our study provides insights for collaborative motion of dyads not seen in other work, and forms the basis for developing a controller capable of handling complex tasks.
\end{sloppypar}

\subsection{Experimental Setup}
After obtaining IRB approval, we set up trials involving 2-person teams or dyads in a leader-follower setup. These teams were to work together to perform a series of 6 object-manipulation tasks. For half the experiment, the follower was blindfolded, so all communication had to occur haptically.

\subsubsection{Table}
The object the teams moved was a 59x122x2 cm wooden board -- meant to simulate an object (like a table) that is difficult for one person to maneuver. Attached to the leader end of the board were a pair of ABS 3D-printed handles, to which two ATI Mini45 force/torque sensors were fastened. The sensors transmitted data via ATI NET F/T Net Boxes, which passed data over Ethernet to the computer at a rate of 100 Hz. 

\begin{figure}[t]
  \centering
  \includegraphics[width=0.9\linewidth]{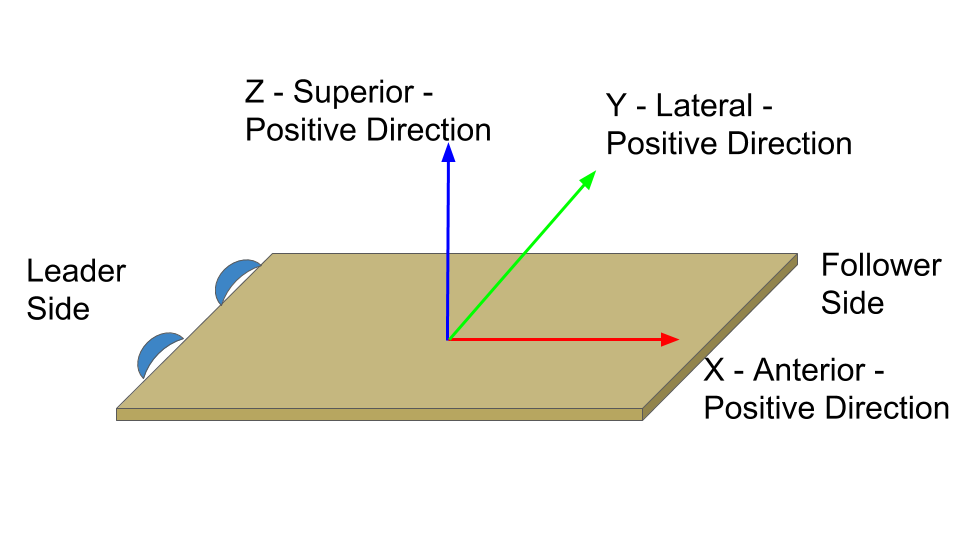}
  \caption{Anatomical direction reference with corresponding table axis: X is anterior, Y is Lateral, and Z is Superior}
  \label{fig:tabdir}
\end{figure}

\begin{sloppypar}
The position of the board was tracked via Cortex Motion Capture software with a Motion Analysis Kestrel Digital Realtime System. A total of 8 Kestrel cameras were used to track 8 infrared markers placed on the board.  Using a static global frame established by the motion capture system, the position and orientation of the board could be tracked over time, and the force and torque data could be transformed into the board's frame, located at the geometric center of the table (see Fig. \ref{fig:tabdir}), as well as the static global frame. The motion capture data was collected at a rate of 200 Hz.
\end{sloppypar}

\begin{figure}[t!]
\captionsetup[subfigure]{justification=centering}
\centering
\subfloat[Starting position for all tasks]{
    \includegraphics[width=.95\linewidth]{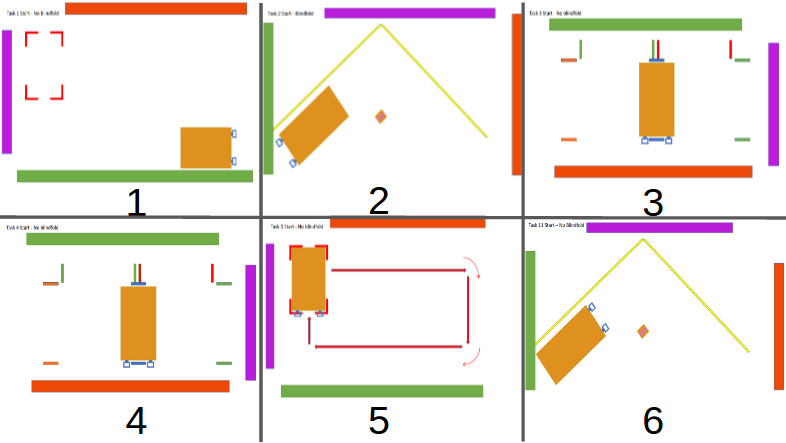}
    \label{fig:AllStart}
}\\
\subfloat[Ending position for all tasks]{
    \includegraphics[width=.95\linewidth]{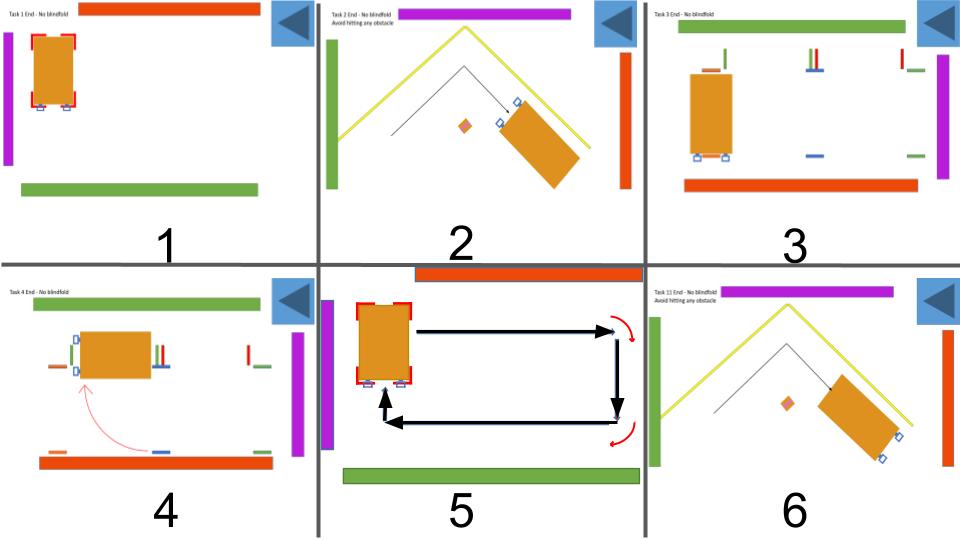}
    \label{fig:AllEnd}
}
\caption{Tablet views of all task instructions for experiment}
\end{figure}

Along with the infrared markers and force/torque sensors, the board also held an Ethernet switch, a power strip, and all cables necessary for power and communication. One experimenter was tasked with making sure no obstacles would trip the subjects, including moving these cables as necessary without exerting forces on the table. During the trials, a tablet was mounted on the board to display instructions to the leader. In total, the board weighed 10.3 kg. An annotated visual of the board can be seen in Fig. \ref{fig:TableConfig}.

\subsubsection{Subjects}
The trial participants were outfitted with polyester arm sleeves for both arms. Two groups of four infrared markers were placed on rigid plates, and then attached to the sleeve, one on the upper arm and one on the lower arm. A blindfold was also used for the tasks where no verbal or visual communication was allowed.

\subsubsection{Arena}
\begin{figure}[t]
  \centering
  \includegraphics[width=0.7\linewidth]{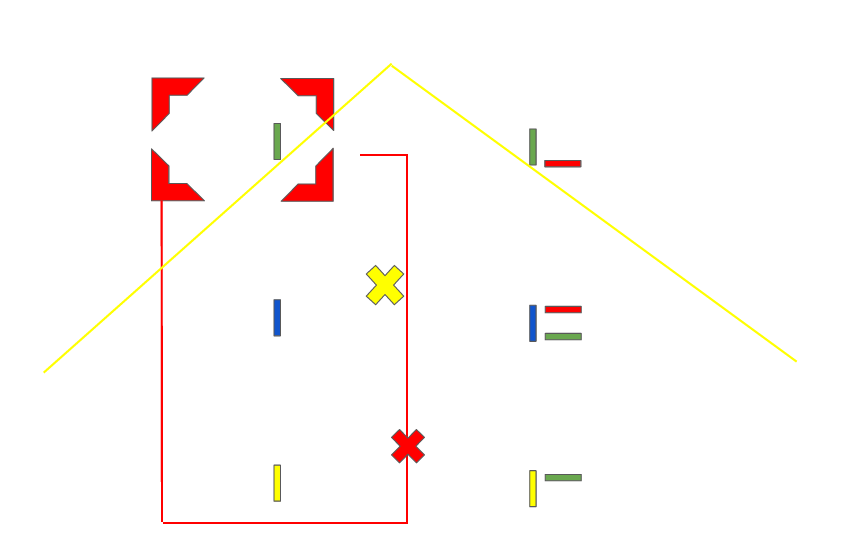}
  \caption{Colored tape used for task delineation}
  \label{fig:TaskDelineation}
\end{figure}

The test arena was a volume measuring 490x510x250 cm. A series of colored tape lines (see Fig.~\ref{fig:TaskDelineation}) were placed on the floor of the volume, indicating key positions for each of the 6 object-manipulation tasks. On 3 of the walls surrounding the arena, we placed green, orange, and purple poster boards to help orient the leader when looking at the tablet. As seen in Fig. \ref{fig:AllStart} and Fig. \ref{fig:AllEnd}, there are colored bars on the edges of each task figure representing the walls with the corresponding color. This way, the leader could more easily determine the frame of reference for the instructions on the tablet mounted to the table.

The arena was also equipped with a video capturing device. The device we used was a Microsoft Kinect 2, which allowed us to capture 3D point cloud data, as well as color video of each trial. Although we did not use the point cloud data for analysis in this paper, the data may be useful in future work.

\subsection{Experimental Procedure}

First, the participants were oriented on the purpose of the research and signed release forms. Second, a leader was chosen at random (by coin flip). Third, each participant put on the sleeves and the participant designated as the follower placed the blindfold on their head, but not covering their eyes until they were about to perform a blindfolded task. Fourth, two preliminary test runs were performed by the participants with the researchers supervising. These test runs walked the participants through each motion required by the tests -- that is translation in x,y, and z axes and rotation in x,y, and z axes (see Fig. \ref{fig:tabdir} for directions). The first run was done without the follower blindfolded, and the second was with the follower blindfolded. Fifth, the leader then was oriented on following the task instructions via the tablet on the table (see Fig. \ref{fig:AllStart} and Fig. \ref{fig:AllEnd}). The researchers displayed the task with visual instructions on the tablet, which corresponded to the colored tape on the ground. The leader then followed the instructions as outlined. Sixth, each group ran through the tasks for 1 hour, which allowed each group to run all 6 tasks approximately 6 times each. The tasks were split evenly between blindfolded and non-blindfolded, and were randomized in order for each group of participants. For instance, a group might perform task 1 non-blindfolded, followed by task 4 blindfolded, followed by 3 blindfolded, and so on. We recognized that learning of tasks would occur, but decided the randomization of task order would help to reduce the amount of guessing of the follower, and would encourage a reactive, rather than anticipatory, response. A researcher changed the setup between tasks, and two other researchers ran data collection for motion capture, force/torque, and video. Finally, the participants were debriefed, they filled out a questionnaire about the trials, and were paid.

The tasks were designed in order to mimic standard motions that humans use when collaborating on moving an object (see Fig. \ref{fig:AllStart} and Fig. \ref{fig:AllEnd} for reference), and are outlined as follows: 
\begin{enumerate}
\item Pick and Place
\begin{itemize}
\item Translation and rotation, but emphasizing the location and orientation of object placement
\end{itemize}
\item Rotation and Translation -- Leader facing backwards
\begin{itemize}
\item Rotation and translation as needed to navigate trial, meant to simulate a narrow hallway, requires rotation about 2 axes
\end{itemize}
\item Pure Translation
\begin{itemize}
\item Translation in the lateral direction
\end{itemize}
\item Pure Rotation
\begin{itemize}
\item Rotation about the superior axis
\end{itemize}
\item 3D Complex Task -- Translation and Rotation in multiple axes
\begin{itemize}
\item Translation in all three axes while avoiding certain 3D obstacles
\end{itemize}
\item Rotation and Translation -- Leader facing forwards
\begin{itemize}
\item Rotation and translation as needed to navigate trial, meant to simulate a narrow hallway, requires rotation about 2 axes
\end{itemize}
\end{enumerate}

\begin{sloppypar}
The physical execution of the task started with each participant grasping an end of the board, the leader by the end with sensors and the follower by the end without sensors. They would then lift the table. After which the follower tried to follow the leader as the leader performed the task indicated on the tablet. Once they reached the position, they set the board back on the ground and released. This constituted a single trial. During sighted trials the participants were allowed any method of communication desired. Whereas during blind trials the participants were only allowed to communicate via forces applied to the board. A sample of task 5 being performed blindfolded and not blindfolded can be seen at \url{https://youtu.be/i-s1pIs17oY}. This task is shown as it encapsulates a majority of the motions seen in all the tasks as they were performed.
\end{sloppypar}

\subsection{Data Collection}
A total of 21 groups participated, and subjects for the trials were recruited using fliers, social media, and word-of-mouth. Trials occurred during February and March of 2016. The participants were comprised of 26 men and 16 women of ages 18-38, and the average age was 22. There were 38 right-handed and 4 left-handed. A scheduling website was used to facilitate trial sessions, and participants signed up for an available hour-long slot.

If, during a task, any error occurred -- such as participants performing a task incorrectly or a failure in data collection -- the task was to be stopped and repeated.

\subsection{Data Analysis}
As previously stated, the data acquired for each trial was the force and torque data from the sensors on each handle, the position and orientation of the table, the position and orientation of the participant's arms, as well as the point cloud data from the Kinect 2. The data we were most interested in initially was the force and torque data in relation to the position and orientation of the table. Data from the blind trials was captured with the objective of characterizing force patterns in human-human dyads that could be used 1) as a baseline for when human-human dyads only use haptic information for co-manipulation and 2) to eventually create a co-manipulation controller that incorporates the force patterns discovered in analysis. Sighted data was captured with the objective of comparing the performance of future co-manipulation controllers with an unrestricted human-human dyad.

\section{Observations}\label{obsv}

\begin{sloppypar}
Although the experiment involved 6 different tasks with up to 6 DOF, this paper focuses on determining a control strategy for 3 DOF planar motion. Since nearly all previous methods involve 1 or 2 DOF, 3 DOF planar motion is a natural step toward 6 DOF control. Because we are focusing on 3 DOF planar motion, our evaluation of the human-human dyad study focuses mainly on the blind versions of tasks 3 and 4, which involved aspects of planar 3 DOF motion. The emphasis was placed on these tasks for a few of reasons. First, as discussed in Section \ref{related_Work}, most research done in this area of pHRI co-manipulation involved either lateral movement with no extended object, or only anterior direction movements (see Fig. \ref{fig:tabdir} for directions reference). When co-manipulating an extended object, the intent of the leader is complicated by the rotation-translation problem described in Section \ref{related_Work}. Therefore, characterizing how humans are able to recognize a desired lateral movement with an extended object and distinguish it from a desired rotational movement is key for successful co-manipulation of extended objects. Second, other tasks--such as task 4 and task 5--include components of lateral translation. Therefore knowing the defining characteristics of only lateral motion helps to recognize it in more complex tasks. We also only analyzed the tasks where users were blindfolded to simplify the analysis, since these tasks involved only haptic communication. The sighted tasks will be used as an upper bound of performance.
\end{sloppypar}

\subsection{Interaction Forces} \label{int_force}

\begin{sloppypar}
Interaction forces are the forces that do not directly relate to motion, i.e. the forces applied by each participant that do not accelerate the object. As suggested by Noohi et al. \cite{Noohi2016}, interaction forces could be used as a source of communication. In our study, the force/torque sensors measured the total force applied to the object. From the measurements alone, we cannot discern between external forces -- forces that accelerate the object -- and interaction forces. Therefore, we could only calculate the interaction force after the experiment ended.

Eq. \ref{ftot} shows the combined forces that contribute to the total force, or the force measured by the sensors. $F_t$ is the total force on the object, $F_i$ is the interaction force, and $F_e$ is the external force causing the object to accelerate. The motion capture data we recorded described the pose of the table over time, and was differentiated twice to acquire the acceleration data. With a known mass of the table and acceleration, the external force was estimated (Eq. \ref{fext}), and removed from the total force to give us the interaction force for each task. 
\end{sloppypar}

\begin{equation}\label{ftot}
F_t = F_i + F_e
\end{equation}
\vspace{-10px}
\begin{equation}\label{fext}
F_e = ma
\end{equation}

For the anterior, $X$, and lateral, $Y$, directions, the only external force being applied is the force applied from the participants, whereas in the vertical $Z$ direction, gravity also caused a force. For all calculations and analysis in this paper, the forces were low-pass filtered near 20 Hz to represent human response ranges. The muscle response of humans can reach up to 100 Hz for brief, forceful efforts, but often lies within the 10-30 Hz range \cite{Burdet}.

Additionally, we calculated the resultant torque on the object at the center of mass. We assumed the table was a rectangular prism with width $w$, length $l$, and depth $d$. We then calculated the torques using Eq. \ref{eq:torque_on_board}, assuming the coordinate frame shown in Fig. \ref{fig:tabdir}. For this calculation, the torques measured by the sensors were neglected since they significantly smaller than the torque due to the reaction forces. $F_{tl}$ and $F_{tr}$ are the total forces measured by the left and right force/torque sensors. Using Eq. \ref{eq:torque_eq} and again getting angular acceleration and velocity from the motion capture data, we calculated the torque causing angular acceleration, getting moments of inertia for a rectangular prism, and assuming planar motion. Finally, we calculated the interaction torque with Eq. \ref{eq:calc_int_tq}.

\begin{equation} \label{eq:torque_on_board}
\begin{aligned}
    \tau_{t,x} = (F_{tl,z} - F_{tr,z})\frac{w}{2} + (F_{tr,y} + F_{tl,y})d \\
    \tau_{t,y} = (F_{tr,z} + F_{tl,z})\frac{l}{2} - (F_{tr,x} + F_{tr,y})d \\
    \tau_{t,z} = (F_{tr,x} - F_{tl,x})\frac{w}{2} - (F_{tr,y} + F_{tl,y})\frac{l}{2}
\end{aligned}
\end{equation}

\begin{equation} \label{eq:torque_eq}
\begin{aligned}
    \tau_{e,x} = I_{xx}\alpha_x - (I_{yy} - I_{zz})\omega_y\omega_z \\
    \tau_{e,y} = I_{yy}\alpha_y - (I_{zz} - I_{xx})\omega_x\omega_z \\
    \tau_{e,z} = I_{zz}\alpha_z - (I_{xx} - I_{yy})\omega_x\omega_y 
\end{aligned}
\end{equation}

\begin{equation} \label{eq:calc_int_tq}
\begin{aligned}
\tau_t = \tau_i + \tau_e
\end{aligned}
\end{equation}

\begin{figure}[tbh] 
\centering 
	\includegraphics[width=0.9\linewidth]{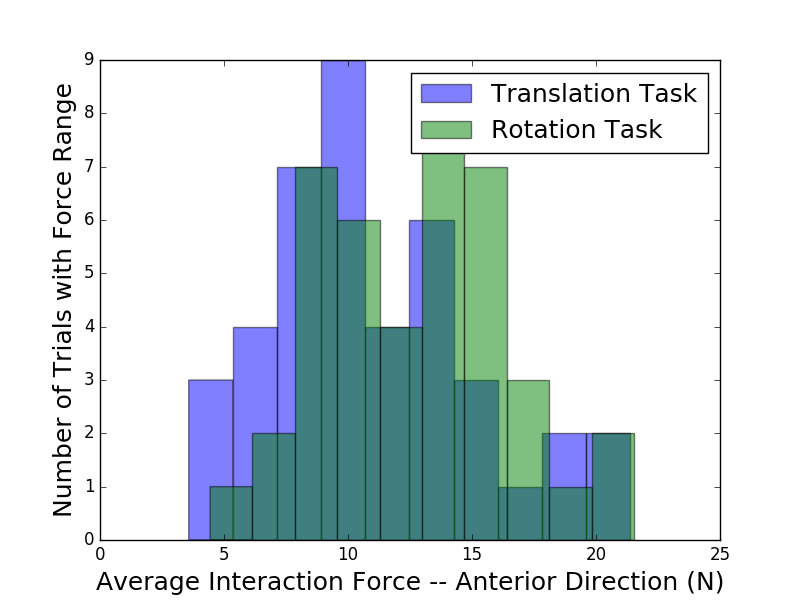}
    \caption{Histogram showing average force along anterior axis for rotation and translation tasks}
    \label{fig:int_force}
\end{figure}

\begin{sloppypar}
As mentioned in Section \ref{related_Work}, some prior work in pHRI has presented an objective of pHRI controllers as minimizing interaction forces by driving them to zero magnitude \cite{Groten2013}. In fact, this is also a characteristic of variable-impedance control \cite{Duchaine2007,Ikeura1965}. Our study, however, showed that this may not always be the case. For both lateral and rotational movements, we calculated the average interaction force in the anterior direction. Histograms showing the distribution of average force  over the duration of trial, for all translation and rotation trials, are shown in Fig. \ref{fig:int_force}. This plot shows a histogram of average interaction force. As can be seen, the interaction force was almost always non-zero for both lateral and rotational movements in the anterior direction (see Fig. \ref{fig:tabdir} for clarity on directions). Additionally, we considered the ratio of average interaction to external forces, $F_{i,avg}/F_{e,avg}$, which--when averaged over all the trials--gave a magnitude of 20, indicating the forces used for acceleration of the object were 20 times smaller than those not used for acceleration. It is not clear why the average interaction force was so substantial, but our hypotheses include: 
\end{sloppypar}

\begin{enumerate}
	\item These forces were used for object and human stability
	\item These forces were used to better communicate intent
\end{enumerate}

We will conduct future studies to explore the hypothesis on stability, but the hypothesis on communicating intent is discussed to some extent in Sections \ref{lmsc} and \ref{evic}. This result is important because it implies that lateral collaborative movements may rely on interaction forces along the anterior direction, which is not seen in many state-of-the-art pHRI controllers. Additionally, minimizing interaction forces may not yield results easily understood by human partners in co-manipulation tasks, since it is now evident that humans are not necessarily minimizing these forces. 
We also ran a Wilcoxon rank-sum test on the distributions in Fig. \ref{fig:int_force} to determine if the tasks were statistically different. The test returned a $p$ value of $0.043$, indicating that the different task types were likely to have come from different populations. This indicates that there is some fundamental difference in the forces applied during these two tasks, and could show how human-human dyads solve the rotation-translation ambiguity problem. This is discussed further in Sections \ref{lmsc} and \ref{evic}.

\subsection{Minimum-Jerk}\label{minjerk}
\begin{figure}[t]
\captionsetup[subfigure]{justification=centering}
\centering
\subfloat[Lateral task, Pearson correlation coefficient is 0.59]{
	\includegraphics[width=0.9\linewidth]{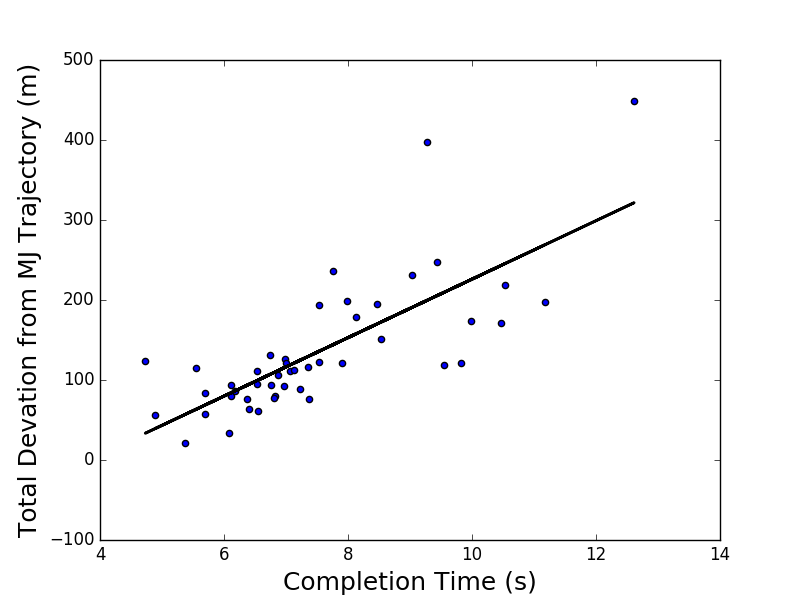}
    \label{fig:metcom}
}\\
\subfloat[Rotation task, Pearson correlation coefficient is 0.83]{
	\includegraphics[width=0.9\linewidth]{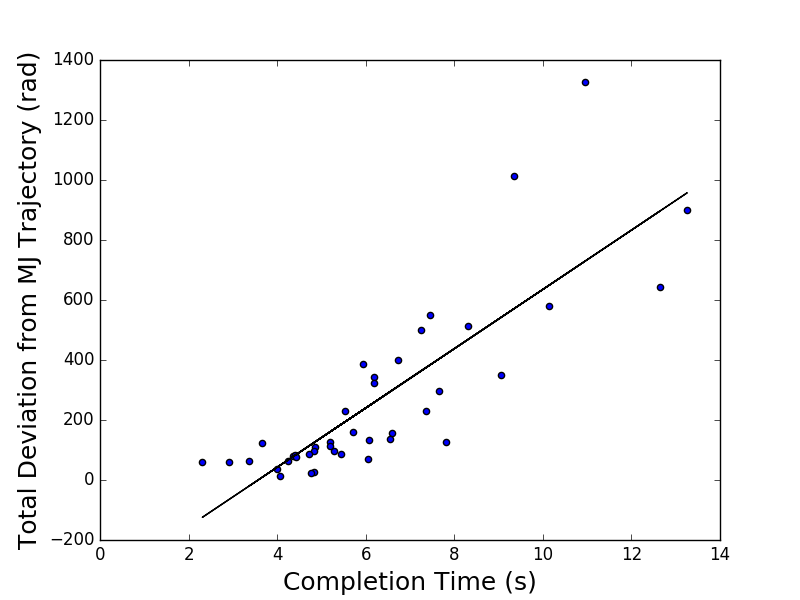}
	\label{fig:ang_mj_comp}
}
\centering
\caption{Comparison of completion time to deviation from MJ trajectory with trend line}
\end{figure}

\begin{figure}[t]
\captionsetup[subfigure]{justification=centering}
\centering
\subfloat[Lateral task trajectories]{
	\includegraphics[width=0.9\linewidth]{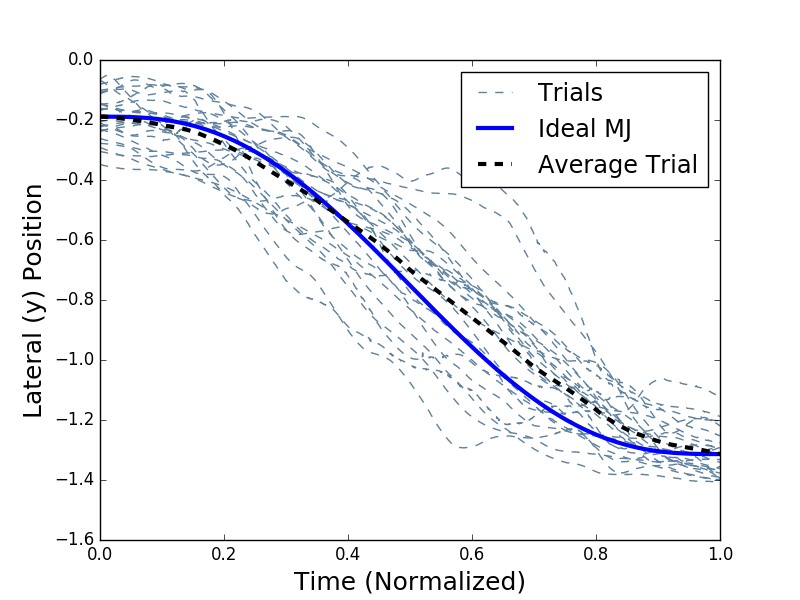}
    \label{fig:negmj}
}\\
\subfloat[Rotational task trajectories]{
	\includegraphics[width=0.9\linewidth]{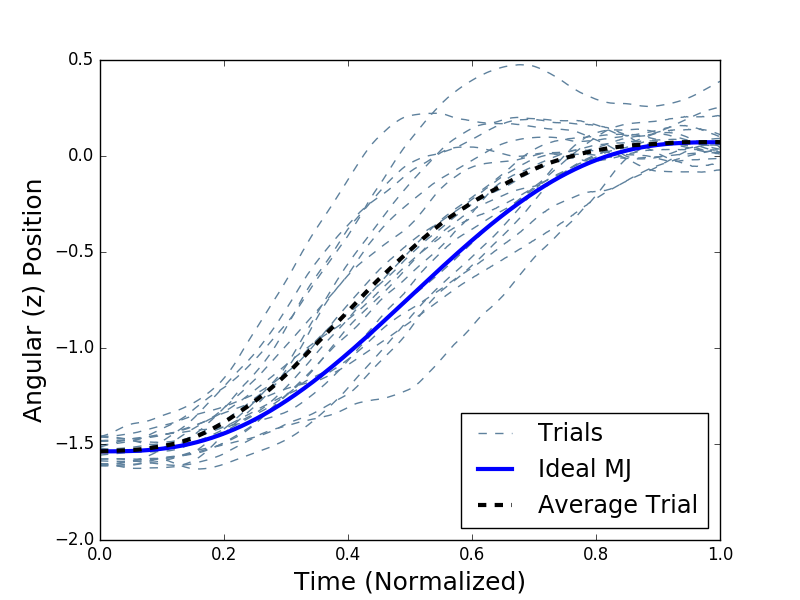}\par
	\label{fig:neg9}
}
\centering
\caption{Individual trial trajectories, ideal MJ trajectory, and average trial trajectory, all scaled to normalized time}
\end{figure}

\begin{sloppypar}

The minimum-jerk (MJ) movement is well-documented as a basis for human arm movements, especially in point-to-point movements. However, we did not expect to see MJ trajectories in these trials, since one participant was blindfolded and unaware of the task specifications, and the dyads used whole-body motion rather than arm-only motion. However, another interesting finding from our study was that both the lateral movement tasks and the planar rotation tasks resembled a MJ movement in lateral position and angular position respectively. This was especially true for the dyads that completed the task more quickly. Figs. \ref{fig:metcom} and \ref{fig:ang_mj_comp} show a positive correlation between deviation from MJ trajectories and time to complete the task. The Pearson correlation coefficient for Figs. \ref{fig:metcom} and \ref{fig:ang_mj_comp} was 0.59 and 0.83 respectively. Deviation from MJ trajectories, or MJ Error, is simply the sum of the residuals between the actual trajectory and ideal MJ trajectory. When considering task completion time, the slower dyads often had a larger error between their position, and the ideal MJ position, whereas the quicker dyads generally had a smaller error with respect to a MJ trajectory. This could be due to the follower becoming confused, and taking more time to determine the intent of the leader. Therefore, completion time or deviation from a MJ trajectory could be an indication of how much confusion was present in the dyad.

Overall, the lateral position and angular position stayed close to the MJ trajectory, and adhering to a similar trajectory over all trials corroborates the results of similar 1-dimensional studies \cite{Bussy2012a}. Figs. \ref{fig:negmj} and \ref{fig:neg9} show the position of the tasks over a normalized time, since each trial took a different amount of time. The gray dotted lines show each individual task, the black dotted line is the average position of all the tasks, and the blue line is the ideal MJ trajectory given an average start and stop position. As we can see, even though the follower did not know the end position, they managed to remain fairly close to the MJ trajectory both for translation and rotation tasks. The ideal MJ trajectory is calculated using Eq. \ref{eq:calc_min_jerk}, where $\hat{t} = t/(t_f - t_0)$, and $t_0 $ is the starting time and $t_f$ is the ending time.

\end{sloppypar}

\begin{equation} \label{eq:calc_min_jerk}
\begin{aligned}
x(t) = x_0 + (x_f - x_0)(10\hat{t}^3 - 15\hat{t}^4 + 6\hat{t}^5) \\
\end{aligned}
\end{equation}

\begin{sloppypar}

Despite the evidence presented for planar co-manipulation tasks, it is not clear at this point if MJ trajectories encompass general co-manipulation for 6D tasks. Dyads may have confusion about intent, may encounter obstacles, or may move for indefinite amounts of time. These situations can lead to non-MJ trajectories, as we have seen in our study (see Section \ref{lmsc}), and we agree with previous research that a MJ basis for control may be too restrictive \cite{Thobbi2011}. However, we also conclude that MJ trajectories can be useful for describing task metrics (discussed more in Section \ref{metrics}), as was proposed in other work \cite{Ivaldi2012}.

\end{sloppypar}

\subsection{Metric Observations}\label{metrics}

\begin{sloppypar}
Our objective for the future is to develop control algorithms for a human-robot dyad to successfully co-manipulate extended objects in 6 DOF tasks. We have begun some initial work on human-intent estimation from the data we collected and a demonstration of simple pHRI \cite{Townsend}. However, to compare a co-manipulation controller with human-human dyads, we need to first be able to quantify the performance of human-human dyads. Therefore, we needed to identify and define which metrics characterize human-human performance. There are a few difficulties in determining performance metrics from our experiment. First, the dyads were given no specific directions other than for the leader to complete the on-screen tasks, and the follower to follow the leader. Therefore, we cannot say a specific dyad was intending to complete the task quickly, or precisely, rather, we can only say that the dyads performed the task in whatever way they preferred. Second, metrics may not be universal across all tasks. For instance, following a minimum-jerk trajectory could be a potentially good metric for translation tasks, but not necessarily for the complex 3D task in our human-human trials. 

Given these difficulties, we sought to find behaviors and trends among the dyads, such as those in Section \ref{minjerk}, that could be used to indicate default or nominal human performance. These metrics indicate the average behavior of human-human dyads, such as how humans have been shown to innately fall into MJ patterns in reaching movements \cite{Flash1985}. We also desired to look into classical metrics that indicate the effective performance of a dyad, such as completion time, power, wasted energy, etc. Our reasoning in looking at these two types of metrics is that we can quantify what people do on average. Although useful in making a robot intuitive, those nominal human-human metrics may actually contradict other effective performance metrics for improving things like energy expended or completion time.
\end{sloppypar}

With this information in mind, we analyzed our data for metrics to classify both the nominal and effective performance of the dyads. Some of the metrics considered were:

\begin{itemize}
	\item Task completion time
	\item Average/max force and torque on table
	\item Average/max velocity of table
	\item Average/max angular velocity of table
	\item Deviation from MJ trajectory of table
	\item Average/total torque change
\end{itemize}

\bgroup
\def\arraystretch{1.25}
\begin{table}[t]
	\fontsize{11}{9}
	\caption{Metric correlation for translation task.}
	\label{table:trans}
\begin{center}
	\begin{tabular}{| c | c | c | c | c | c |}
	\hline
	& MJ Err & $t_c$ & $v_{y,avg}$ & $\omega_{z,avg}$ & $\Delta\dot{\tau_z}$\\
	\hline
	MJ Err & - & 0.63 & -0.42 & 0.05 & -0.02 \\
	\hline
	$t_c$ & - & - & -0.66 & -0.40 & 0.02 \\
	\hline
	$v_{y,avg}$ & - & - & - & 0.30 & -0.07 \\
	\hline
	$\omega_{z,avg}$ & - & - & - & - & 0.04 \\
	\hline
	\end{tabular}
\end{center}
\end{table}
\egroup

\begin{sloppypar}
We evaluated more than just these metrics, but chose a representative set for our analysis here. The chosen set was compared using the Pearson correlation coefficient, and the results are summarized in Table \ref{table:trans} for the translation task and in Table \ref{table:rot} for the rotation task. The metrics used in these tables are MJ error, defined in Section \ref{minjerk}, completion time, average lateral velocity, average angular velocity, and torque change. Torque change, was defined in \cite{Ivaldi2012}, and is calculated using $\Delta\dot{\tau} = \int_{0}^{T}\dot{\tau_1}^2 + \dot{\tau_2}^2 dt$. 
\end{sloppypar}

\bgroup
\def\arraystretch{1.25}
\begin{table}[t]
	\fontsize{11}{9}
	\caption{Metric correlation for rotation task.}
	\label{table:rot}
\begin{center}
	\begin{tabular}{| c | c | c | c | c | c |}
	\hline
	& MJ Err & $t_c$ & $v_{y,avg}$ & $\omega_{z,avg}$ & $\Delta\dot{\tau_z}$\\
	\hline
	MJ Err & - & 0.83 & -0.50 & -0.56 & 0.71 \\
	\hline
	$t_c$ & - & - & -0.51 & -0.78 & 0.52 \\
	\hline
	$v_{y,avg}$  & - & - & - & 0.81 & -0.36 \\
	\hline
	$\omega_{z,avg}$ & - & - & - & - & -0.42 \\
	\hline
	\end{tabular}
\end{center}
\end{table}
\egroup

We expected there to be some correlation between most of these measurements since most of the metrics were proposed in previous research \cite{Thobbi2011,Miossec2008,Groten2011,Ivaldi2012}. Surprisingly, some intuitively related metrics offered very little correlation. The most relevant expected metrics--for the lateral translation task--were average/max angular velocity and deviation from the MJ trajectory. We expected dyads performing these tasks to minimize the average angular velocity about the $z$ axis, and stay relatively close to the MJ trajectory, but the correlation coefficient between these two metrics was 0.05. In fact, there was a much stronger correlation between deviation from MJ trajectory and completion time and average lateral velocity -- being 0.63 and -0.42 respectively. Intuitively, minimizing angular velocity would be an ideal metric for this task, since a perfect lateral translation would involve no angular velocity at all. This supports our hypothesis that some metrics may be ideal from an energy or time efficiency point of view, but may not be nominal behavior for human users in a dyad, and we need to consider these non-intuitive relationships between metrics if we want to make intuitive robot controllers for co-manipulation.

\begin{figure}[t]
    \captionsetup[subfigure]{justification=centering}
    \centering
    \subfloat[z-axis torque patterns]{
	    \includegraphics[width=0.9\linewidth]{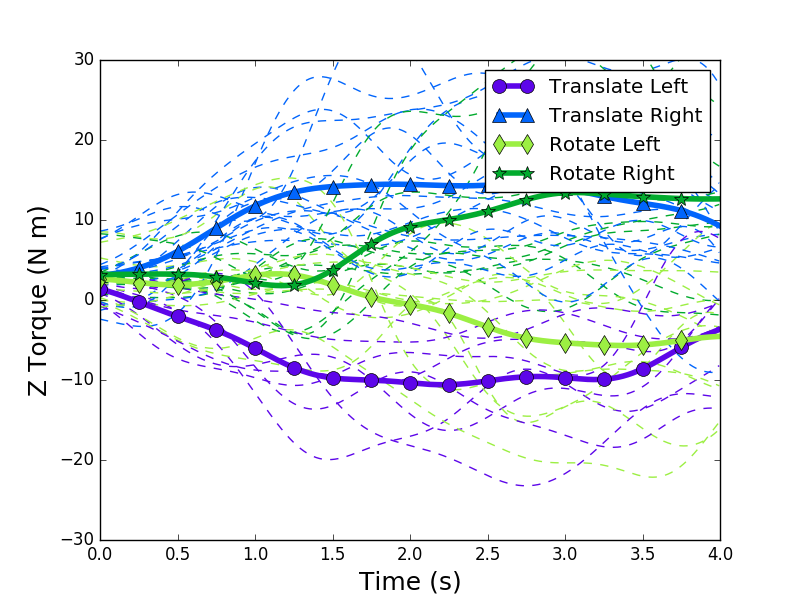}
	    \label{fig:ztorqs}
    }\\
    \subfloat[x-axis torque patterns]{
        \includegraphics[width=0.9\linewidth]{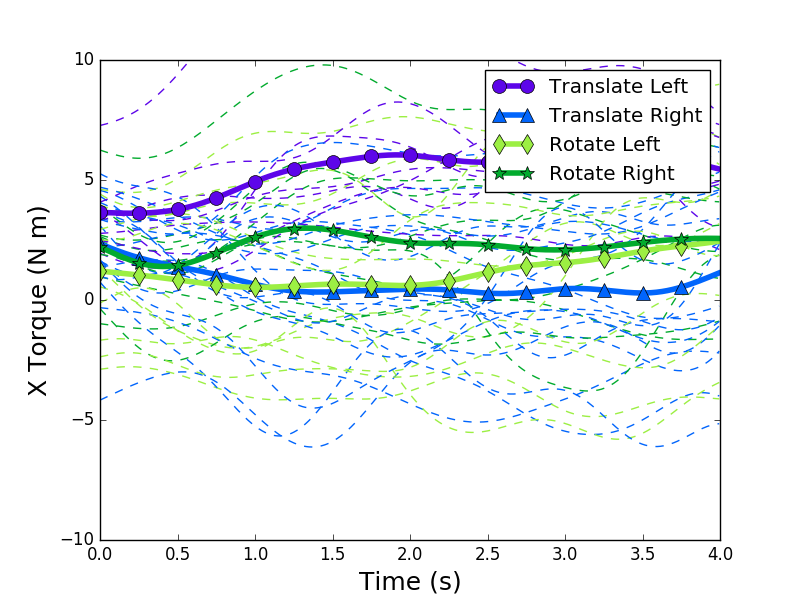}
        \label{fig:xtorqs}
    }
\centering
\caption{First 4 seconds of trials showing torque trends for rotation and translation tasks for both directions of motion: dashed lines are individual trials, bold lines are average over all types of trials}
\end{figure}

Another noteworthy observation is the lack of correlation between total torque change and the other metrics. We would expect that torque change would indicate some measure of performance, but it appears to be more related, at least in the translation case, to some other aspect of the movement. This may be related to our findings from Section \ref{int_force}, where there were interaction forces in the anterior direction for lateral movements, and may indicate that the non-correlated metrics may affect stability, communication, or something related to preference which we did not measure. 

For rotational tasks, the same metric set was applied as can be seen in Table \ref{table:rot}. However, we were again surprised by the results. We expected task completion time and MJ error to be related, since we had seen that previously. What we did not expect was torque change to be so heavily correlated to the other metrics, where it was not in the translation case. This finding was significant to us, and indicated that there is some fundamental difference between $\tau_z$ in these two tasks. We explore this phenomenon further in Section \ref{lmsc}.

\begin{sloppypar}
Some additional information on the metrics can be found in Table \ref{table:avg_std}. This table includes the mean and standard deviation for all the metrics we considered here. Some values here will be used to compare human-robot dyads (see Section \ref{evic}). As can be seen the standard of both MJ error and torque change are quite large compared to their mean, whereas the standard deviation of completion time, average velocity, and average angular velocity are not. This indicates that there are multiple different approaches that the human-human dyads used to achieve similar results. In other words, although each dyad may have applied their own force patterns--or taken a different trajectory--the overall behavior, especially average velocity and angular velocity, was similar.

The surprising correlations between metrics shows we do not fully understand how human-human dyads operate. We may be able to conclude that in some instances, humans may lower or raise performance on an effective metric in order to raise or lower performance on an nominal metric, such as how the dyads appeared to raise the energy-increasing interaction forces on lateral translation trials. While these observations provide some insight to this topic, our work on metrics is still an open question that is necessary to explore in order to better design, characterize, and evaluate performance of human-robot co-manipulation controllers.
\end{sloppypar}

\bgroup
\def\arraystretch{1.25}
\begin{table}[t]
	\fontsize{11}{9}
	\caption{Average and standard deviation of metrics for blind rotation and translation tasks.}
	\label{table:avg_std}
\begin{center}
	\begin{tabular}{| c | c | c | c | c |}
	\hline
	& \multicolumn{2}{|c|}{$\mu$ (mean)} & \multicolumn{2}{|c|}{$\sigma$ (std. dev.)}\\
	\hline
	& Rot & Trans & Rot & Trans \\
	\hline
	MJ Err & 392.71 & 149.91 & 391.7 & 87.65  \\
	\hline
	$t_c$ & 7.08 & 7.18 & 2.9 & 1.62  \\
	\hline
	$v_{y,avg}$  & 0.17 & 0.18 & 0.06 & 0.04  \\
	\hline
	$\omega_{z,avg}$ & 0.26 & 0.004 & 0.09 & 0.003 \\
	\hline
	$\Delta\dot{\tau_z}$ & 488454.4 & 387937.6 & 560601.9 & 281393.2\\
	\hline
	\end{tabular}
\end{center}
\end{table}
\egroup

\subsection{Lateral and Rotation Movement Characteristics}\label{lmsc}

\begin{sloppypar}
In the case of lateral movements, we recognized some patterns in how the dyads behaved. Studying the videos of the lateral motion task, we saw that the follower often guessed the leader's intent incorrectly, and began to rotate when the leader started their movement. When this happened, the leader would flex their arm on one side of the table, causing a torque on the table, and the follower would then commence moving in the correct manner. With this video evidence, as well as the torque change information from Tables \ref{table:trans} and \ref{table:rot}, we began looking for in-task patterns of applied torques which could indicate the leader's intent to start either a translation or rotation task. 
\end{sloppypar}

In order to see in-task relationships, we decided to take each task and look at the time-series torque for the beginning of the tasks. As we looked at the torque data, we noticed two groups beginning to appear. These two groups represented the torque values for the direction of the rotation task, since the dyads were assigned to randomly rotate either clockwise or counterclockwise for each rotation task performed. We then tried looking at the same z-torque time-series data for the translation tasks, and noticed that two more groups appeared, indicating that there was some pattern showing the difference between translation and rotation tasks, as well as a difference depending on which direction the table was travelling. We then took an average of z-torque for each of the 4 distinct groups: translation left, translation right, rotation clockwise (left), and rotation counterclockwise (right). We noticed there appeared 4 groupings of average z-torque for the entire time series. These findings are summarized in Fig. \ref{fig:ztorqs}. 

As can be seen, translation tasks tend to increase in z-torque more quickly, whereas the rotation tasks hover around the same value for over 1 second before diverging. It is evident from this plot that there is a clear difference in torque patterns between the translation and rotation trials, and also the direction of travel. The intent can be classified as either translation left, or translation right, depending on the z-torque value achieved. However, there is not a difference between z-torque patterns for the first second of left and right rotations. Both directions have an approximately constant torque value for this time segment. This is an important time segment, since it is during this interval that decisions about whether to rotate or translate are made by the follower, and there needs to be some indication given by the leader to signal which direction to go.

For this reason, we also looked into what other signals might be given by the leader to indicate which direction to travel, and to clarify whether to rotate or translate. In watching video of the trial, we also noticed some dyads tended to rotate the board about the anterior (x) axis while performing the tasks. So we did a similar analysis with x-torque values to what humans did with the z-torque values. The results can be seen in Fig. \ref{fig:xtorqs}. As can be seen, similar to Fig. \ref{fig:ztorqs}, there is a divide between left translation and right translation. Additionally, a divide appears between left rotation and right rotation around the 0.75 second mark. In order to determine whether the division was significant, we ran a t-test comparing the left rotation and right rotation x-torque patterns. We found that during the $0.75 - 1.75$ second range the $p$ values ranged from $0.1 - 0.2$. Although this isn't significant on a 95\% confidence interval, we believe it is enough of a distinction for use in control. Additionally, we compared left translation and left rotation with a t-test, and found that during the $0.75 - 1.75$ second time range, $p < 0.05$, indicating that the x-torque values can be used to distinguish between rotation and translation going one direction. Similar results were found comparing right translation and right rotation. This means we can use the z-torque to determine direction of travel, and x-torque to determine type of motion.

\begin{figure}[t]
  \centering
  \includegraphics[width=0.9\linewidth]{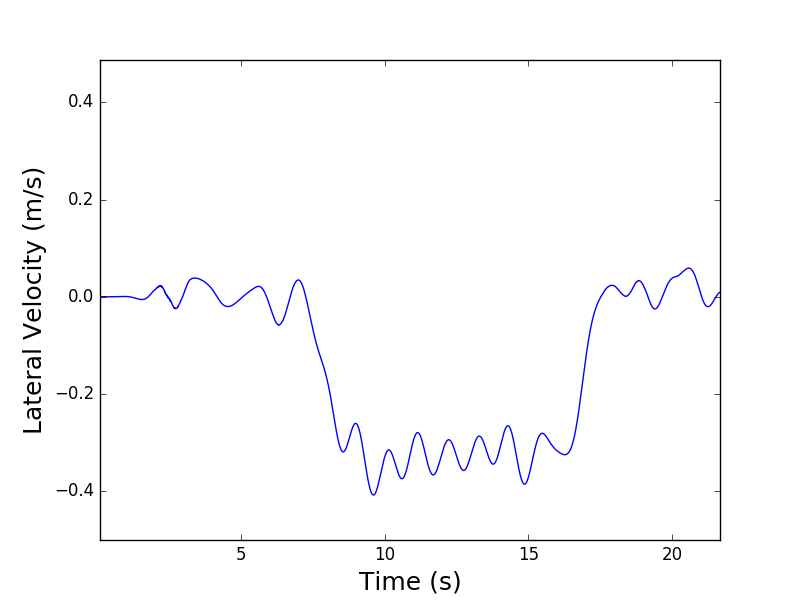}
  \caption{Plot showing lateral velocity profile for beginning of Task 5, a 3D Complex Task avoiding obstacles: this portion of the task is a lateral translation for over 2 meters}
  \label{fig:cvel}
\end{figure}

\begin{sloppypar}
With the analysis on torque triggers providing background for a control method for extended object co-manipulation, we determined what the velocity profile should look like for these tasks. For the translation tasks, we assumed it would follow the bell-shaped velocity profile from a MJ trajectory, however, we wanted to see how the velocity profile looked when translating over a large distance. Bussy et al. \cite{Bussy2012a} showed that humans often accelerate an object to a steady velocity while translating an object. We wanted to verify this, and also determine what velocity most dyads chose as the steady state velocity. To do this, we looked at our 3D Complex Task data. This task involved a large translation portion, followed by changes in direction and rotation of the board to avoid obstacles. Fig. \ref{fig:cvel} shows the first portion of a typical complex task, which is a translation for over 2 meters. We notice from this data that the results seen in Bussy et al. can be verified, and also that the steady velocity achieved is around -0.35 m/s. It is important to note that this velocity value is for a 10.3 kg board, and may differ depending on the mass of the object. Since we want our robot controller to work in an undefined situation, we would expect it to be able to perform a translation task indefinitely, if needed. This observation helps inform us that a dyad's desired velocity is some steady state value.
\end{sloppypar}

\begin{figure}[t]
\captionsetup[subfigure]{justification=centering}
\centering
\subfloat[Control loop for BMVIC]{
    \includegraphics[width=0.98\linewidth]{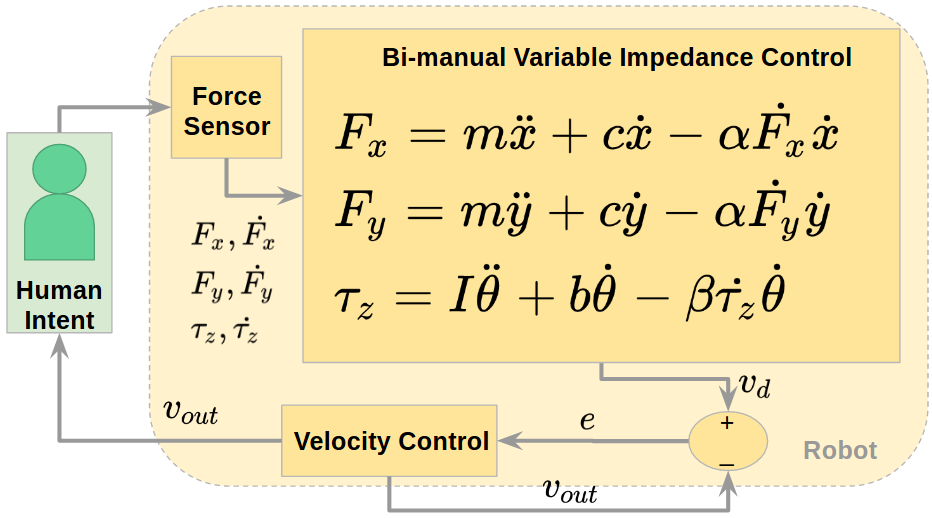}
    \label{fig:hr_loop}
}\\
\subfloat[Control loop for EVIC]{
    \includegraphics[width=0.98\linewidth]{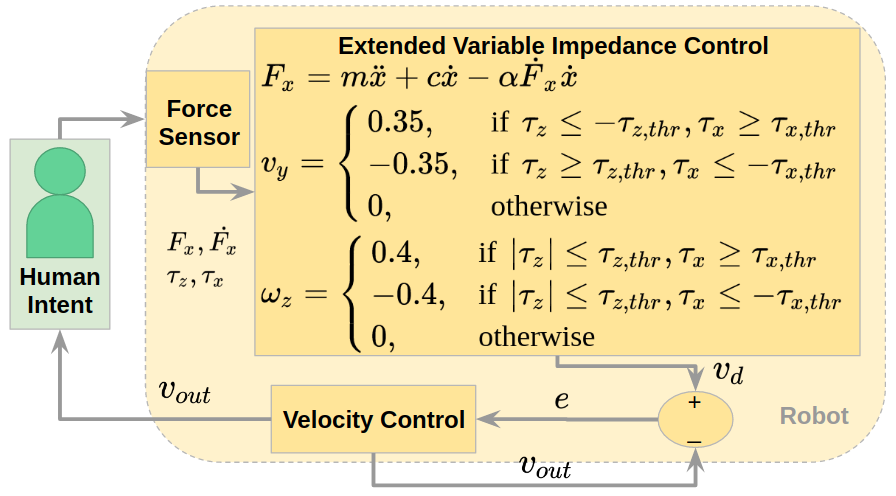}
    \label{fig:alg_pic}
}
\caption{ Control loops for co-manipulation of extended object showing human (left box) communicating intent haptically through force sensor, then desired velocity is calculated using specified control law and sent to velocity controller.}
\end{figure}

\begin{sloppypar}
This new information provides a unique perspective on how force can be used to solve the problem of co-manipulation of extend objects. Using the tendencies of humans in applied torque patterns from the leader, we can distinguish between task type and task direction intent for the follower to use.
\end{sloppypar}

\section{Robot Hardware Platform Description}
\begin{figure}[t]
  \centering
  \includegraphics[width=.75\linewidth]{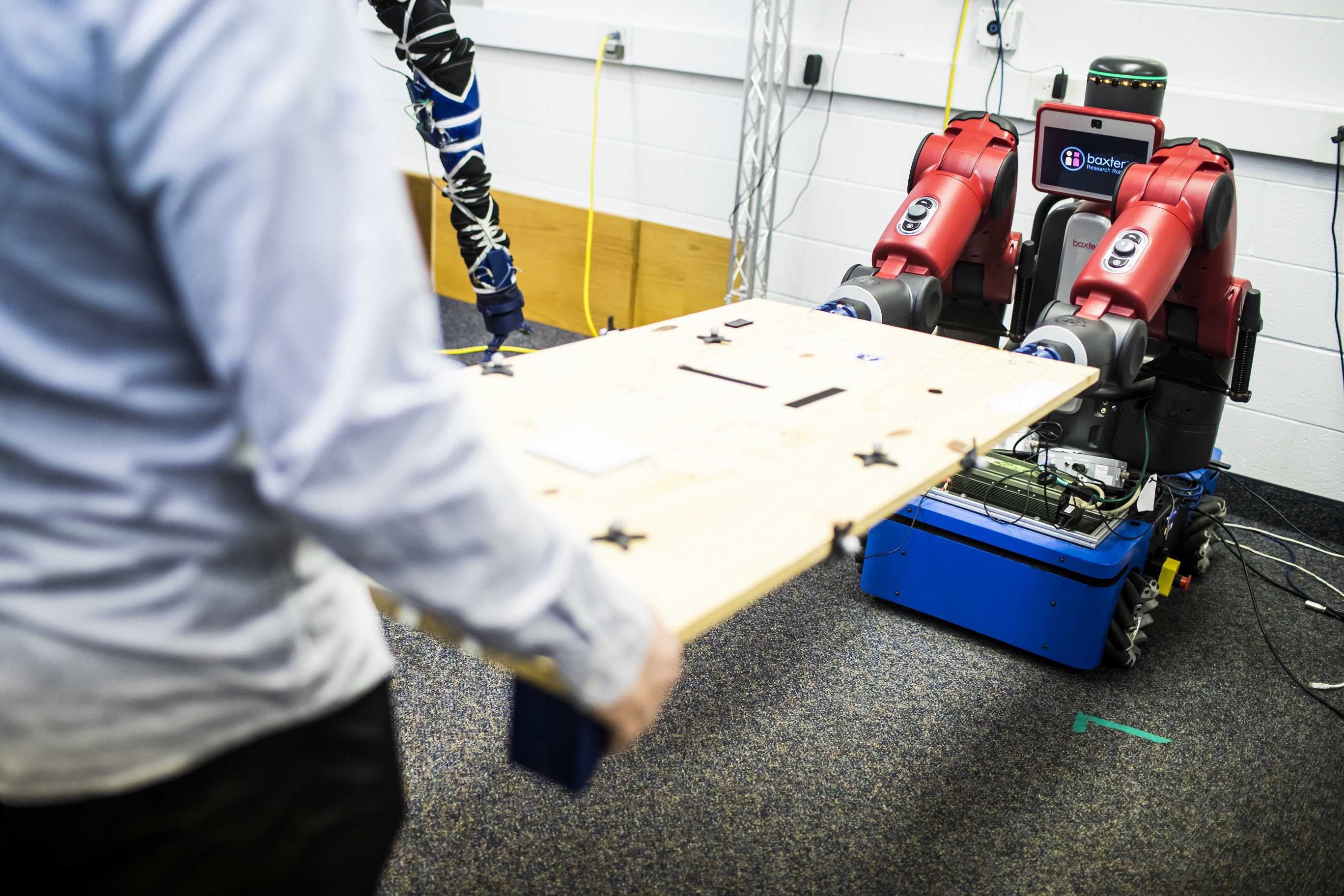}
  \caption{Rethink Robotics Baxter robot mounted on HStar Technologies AMP-1 holonomic base carrying the table with a person.}
  \label{fig:Megazord}
\end{figure}

The purpose of our human-human study and the purpose of creating the physical human-robot interaction methods for co-manipulation defined in Sections \ref{evic} and \ref{sec:neural-network} was to use them for actual co-manipulation. Therefore, we designed an experimental study to determine the capabilities of our controllers for co-manipulation on a real robot platform. Our robot platform for this research is a Rethink Robotics
Baxter robot mounted on an AMP-I holonomic base from HStar
Technologies as seen in Fig. \ref{fig:Megazord}. There are force/torque sensors on Baxter's wrists, and the base is equipped with mecanum wheels. For our initial work, we chose to use a
holonomic base with mecanum wheels instead of a bipedal
robot in order to validate that the human intent prediction works at the appropriate speeds without having to incorporate the complexities of bipedal robots. This is important to ensure that our methods work in real world applications as limiting speed due to limited locomotion may affect the dynamics of the interaction.

For all human-robot experiments described throughout the rest of this paper, the Baxter arms ran an impedance controller with a commanded joint angle calculated for acceptable positioning of the table. The impedance controller was run along with Baxter's built in gravity compensation. The impedance control law, given in Eq. \ref{eq:torque_imp}, used $K_p$ and $K_d$ gains of [40, 120, 40, 16, 8, 10, 12] and [7, 8, 4, 7, 1.5, 1.5, 1] respectively. The same gains were used for both arms. The desired angles, $q_{cmd}$, used were [0, -0.84, -1.27, 2.26, -0.34, -1.22, -2.25] radians and [0, -0.84, 1.27, 2.26, 0.34, -1.22, 2.25] radians for left and right arms respectively. We ran the controller at a rate of 500 Hz.

\begin{equation}\label{eq:torque_imp}
    \tau_{cmd} = K_p(q_{cmd}-q) - K_d\dot{q}
\end{equation}

As described in other literature, \cite{Burdet}, the impedance controller allows the robot to react in a more human-like manner, making the human-robot interaction more intuitive and natural for a human user. This means that for both EVIC and NNPC, the desired velocities are commanded directly to the holonomic base, whereas the arms are not providing any desired motion toward the goal, except through motion due to low impedance. While humans typically use their arms in co-manipulation tasks, especially when doing precise placement, using the impedance control law allows us to run initial studies to determine if our co-manipulation controllers are good approximations for human behavior in co-manipulation.

\section{Planar Extension of Variable Impedance Control} \label{evic}
\subsection{Motivation and Formulation}

\begin{sloppypar}
In order to verify that the torque patterns we saw in Section \ref{lmsc} would be applicable in an extended object co-manipulation scenario, and also to show that current variable impedance co-manipulation techniques in the literature are not adequate for  extended objects, we built an extension onto a variable impedance controller. Variable impedance control (VIC) is a possible solution to undefined or indefinite scenarios, since it is not based on a trajectory, but rather on force inputs which determine robot velocity. What we noticed in practice is that VIC has issues when dealing with bi-manual control and extended object control. We implemented a VIC based on Duchaine and Gosselin's work, \cite{Duchaine2007}, on our robot platform. 
\end{sloppypar}

 Our implementation of VIC, which we called Bi-Manual VIC (BMVIC), involves the control loop seen in Fig. \ref{fig:hr_loop}. The human communicates their intent to the robot through force sensors, and the VIC model determines a desired velocity based on the applied force, and how the force is changing in relation to the robot's velocity. The general model is shown in Eq. \ref{eq:forcevic}, as well as in the figure. Here, $F$ and $\dot{F}$ are applied force and time derivative of force, respectively, $\dot{p}$ and $\ddot{p}$ are velocity and acceleration, and $m$, $c$, and $\alpha$ serve as virtual mass, damping and weighting parameters to define the impedance. These virtual parameters do not correspond to the actual parameters of the system, and have values of $1.2,$ $0.6,$ and $0.2$ respectively, and were determined by trial and error. The model can be discretized and implemented as a discrete LTI system, solving for the desired velocity at each time step. We applied the resulting desired velocity that would give a model impedance directly to the base, and controlled the robot arms to have very low-impedance (see Section \ref{evic_imp}.) This method was developed for single arm manipulation, so we implemented a VIC for each arm independently in order to achieve bi-manual manipulation. However, this is not an ideal method for bi-manual control. Pushing one arm forward and one arm backward would apply zero net force, causing the robot to remain stationary, rather than rotate, as we hoped. To deal with this, we added a torque model to their VIC model, as seen in Eq. \ref{eq:torquevic}. Here, $\tau$ and $\dot{\tau}$ are applied torque and time derivative of torque, respectively, with $\dot{\theta}$ and $\ddot{\theta}$ as angular velocity and acceleration, while $I$, $b$, and $\beta$ serve as virtual inertia, damping, and weighting parameters, with values of $0.12,$ $0.6,$ and $0.2$. All forces and torques referenced here and used for variable impedance control are with respect to the center of the table. The bi-manual torque-based model theoretically allows VIC to be extended to planar motion, where pushing one arm forward and one arm backward will provide a net torque, indicating a desired angular velocity (in the plane only), in addition to any desired Cartesian velocities calculated by the original model. In summary, at each time step, Eqs. \ref{eq:forcevic} and \ref{eq:torquevic} are solved to determine desired velocity and angular velocity to send to the velocity controller. We will refer to the bi-manual torque-based model as bi-manual VIC or BMVIC.

\begin{equation}\label{eq:forcevic}
F = m\ddot{p} + c\dot{p} - \alpha\dot{F}\dot{p}
\end{equation}
\vspace{-15px}
\begin{equation}\label{eq:torquevic}
\tau = I\ddot{\theta} + b\dot{\theta} - \beta\dot{\tau}\dot{\theta}
\end{equation}

\begin{sloppypar}
We also extended VIC in a different and new way, using our results from Section \ref{lmsc}. We used Eq. \ref{eq:forcevic} as a base controller for anterior/posterior translation, and added torque-based triggers for lateral translation and planar rotation. The logic of this extended variable impedance control (EVIC) is shown in Algorithm \ref{alg:evic}. Torque thresholds are calculated, based on Figs. \ref{fig:ztorqs} and \ref{fig:xtorqs}, and are implemented as shown. We centered the thresholds around zero for ease of implementation. The threshold values are $3.0$ $Nm$ for z torque and $1.5$ $Nm$ for x torque. If none of the torque threshold conditions are met, the algorithm commands no lateral translation or rotation about the superior axis. As mentioned previously, if the torque threshold conditions are met, the robot accelerates until it reaches a specified steady state velocity. The lateral velocity value, 0.35 m/s, was determined from the logic described in Section \ref{lmsc} and Fig. \ref{fig:cvel}, and the rotation velocity value, 0.4 rad/s, was determined similarly. The robot acceleration was limited to the capabilities of our robot mobile base. A control loop showing how this algorithm is implemented is shown in Fig. \ref{fig:alg_pic}. The main difference between EVIC and BMVIC is that EVIC uses Algorithm \ref{alg:evic} to determine the desired lateral and angular velocities, whereas BMVIC relies on Eqs. \ref{eq:forcevic} and \ref{eq:torquevic} to calculate the desired lateral and angular velocities.
\end{sloppypar}

\begin{algorithm}[t]
\caption{Extended Variable Impedance Control}\label{alg:evic}
\begin{algorithmic}
\Require $\tau_z,\tau_x$
 \If{$\tau_z\leq-\tau_{z,thresh}$ and $\tau_x\geq\tau_{x,thresh}$}
   \\ \hspace{1cm} Left Translation\;
 \ElsIf{$|\tau_z|\leq\tau_{z,thresh}$ and $\tau_x\geq\tau_{x,thresh}$}
   \\ \hspace{1cm} Right Rotation\;
 \ElsIf{$|\tau_z|\leq\tau_{z,thresh}$ and $\tau_x\leq-\tau_{x,thresh}$}
   \\ \hspace{1cm} Left Rotation\;
 \ElsIf{$\tau_z\geq\tau_{z,thresh}$ and $\tau_x\leq-\tau_{x,thresh}$}
   \\ \hspace{1cm} Right Translation\;
 \Else
   \\ \hspace{1cm} Stop\;
 \EndIf
\end{algorithmic}
\end{algorithm}

This EVIC algorithm represents the average behavior of human-human dyads manipulating a specific object. It is not evident that the thresholds would work for objects of different size or mass. However, in our initial testing of the controller, we used a table of about half the length and mass of the table used in the experiment, and achieved similar general performance of the controller. This generalized behavior, however, was not tested thoroughly. As a general rule, different torque thresholds may need to be set for specific robot platforms as well as different objects. In order to set thresholds for torque, as well as the target velocity, one may consider using a learned approach--or an optimization--where a user would manipulate the object for a certain period of time, and the algorithm would adjust to the preferences of the user and the characteristics of the object, based on the applied forces and achieved velocities. This is beyond the scope of this paper, and may be added in future work.

Also, it is not certain that the force/torque pattern seen in these tasks, nor the torque thresholds used, would be applicable to tasks involving higher DOF. This is something that must be explored in future work as we extend our methods to 6 DOF tasks.

\subsection{Extended Object Co-manipulation Implementation}\label{evic_imp}
We implemented both BMVIC, as well as the EVIC on our robot platform, shown in Fig. \ref{fig:Megazord}. A video showing EVIC running can be seen at \url{https://youtu.be/5vicqv788dI}. Our purpose in implementing these controllers was to determine feasibility of the controllers, and also to get initial data quantifying performance of a human-robot dyad against the blindfolded human-human dyads. As a reminder, BMVIC is a bi-manual implementation of the most relevant pHRI controller found in related literature (see Section \ref{related_Work}) for co-manipulation of an extended object. We ran both controllers testing the capability of planar co-manipulation of an extended object on the following criteria: lateral translation and planar rotation, or rotation about the superior axis. We ran the controller at a rate of 500 Hz, manipulating or carrying the same table from our human-human dyad experiment (see Fig. \ref{fig:TableConfig}.) For determining performance of the controllers, we compared the completion time and MJ error for both lateral and rotational tasks. We also had a qualitative metric: whether BMVIC, EVIC, or neither controller was preferred.

\subsubsection{Pilot Study Testing}

During feasibility testing, we discovered major issues with BMVIC. As stated previously, the issue VIC faces is a problem related to rotation-translation. This problem arises when forces are applied laterally on a long object being manipulated by two partners, and the follower does not know whether the leader wants to rotate or translate. We had hoped that applying Eq. \ref{eq:torquevic} would allow us to overcome the rotation-translation problem. In practice, however, the controller was unable to correctly predict the direction and type of motion desired. Additionally, the robot often became unstable with the human in the loop, and ended up shearing internal components within the arm during two different trials. When running EVIC, these problems were not as prevalent. Incorrect predictions occurred, but only when the user did not move as the algorithm anticipated and this movement did not cause instability. We recognize this does not allow for a detailed comparison between BMVIC and EVIC. But due to the resulting damage on our robot platforms, we instead decided to compare EVIC to human-human data from our study and to the neural-net based controller described in Section \ref{sec:neural-network}. 

\section{Neural Network Control}
\label{sec:neural-network}
\begin{figure}[t]
  \centering
  \includegraphics[width=.9\linewidth]{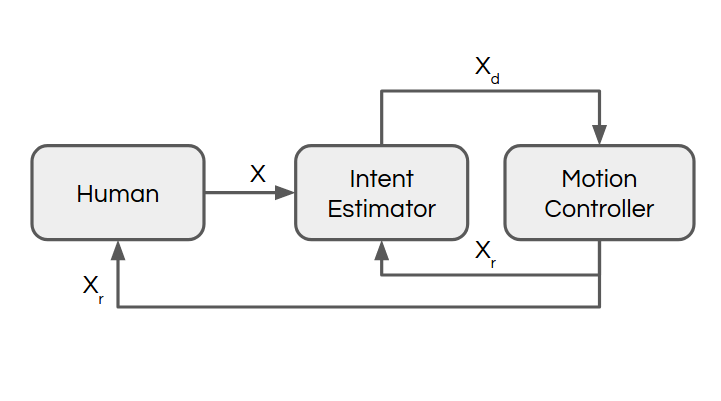}
  \caption{Basic control loop structure of intent estimation in co-manipulation. The human moves the co-manipulated object, and the motion of the object, $x$, is fed into an intent estimator, which determines a desired motion of the robot, $x_d$. The robot's motion, $x_r$ then influences the object motion, as well as the human's.}
  \label{fig:control_loop}
\end{figure}

As an approach to developing a nonlinear estimator of human
intention, we formulated a neural network using the Google TensorFlow
API. As we stated previously, in creating an estimator for co-manipulation control, the objective is to determine how the robot should respond to human intent, i.e. how it should move to achieve the human partner's goal. The intent estimator should allow us to control the object with a control loop similar to that seen in Fig. \ref{fig:control_loop}. 

Because our data considered the interaction between a human leader and a human follower, the input $x$, could be considered what the leader did--applied forces or moved the object--to indicate their intent to the follower. The follower then deciphered the intent, $x_d$, and moved as they believed appropriate, $x_r$. The leader then reacted to this motion. Essentially, the leader put in a stream of data that indicated intent to move the object in a certain manner, which the follower then deciphered and acted on the intent. The estimator should follow this basic input data to velocity command outline.

There are a variety of neural network structures that could be considered for this purpose. Martens et al. showed how Recurrent Neural Networks (RNNs) are used in predicting text. A sequence of characters can be fed into the RNN as an input, and the RNN will predict the next character in the sequence \cite{Martens2011}. In considering what form our neural network should take, this model proved to be the most applicable to our work. From our exploratory study, we had sequences of forces applied to and motion of a table that could be used as inputs to an RNN. We determined that we could feed the force and motion data, as a substitute for characters in other RNNs, and receive a motion prediction as an output, similar to how RNNs are used for predicting text. This prediction encapsulates the human intent, encoded in the desired velocity, and provides a goal for the robot to achieve.

\subsection{Architecture}
\begin{figure}[t]
  \centering
  \includegraphics[width=1.0\linewidth]{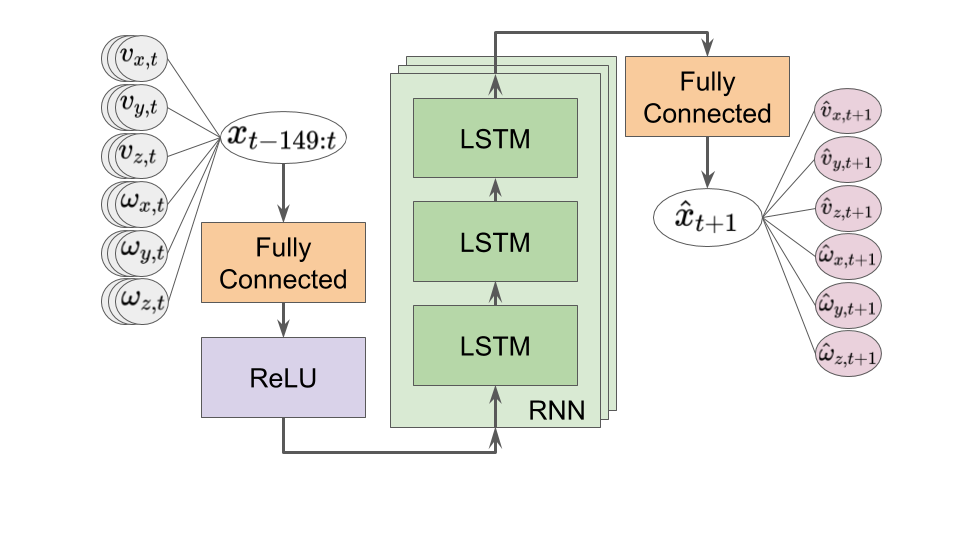}
  \caption{Basic neural network structure, time-series motion data inputs (left) enter the network, are sent through a fully connected layer, a ReLU layer, through an LSTM Cell RNN, and another fully connected layer before predicted velocities are given as outputs (right).}
  \label{fig:structure}
\end{figure}

Once we decided on using the RNN structure, we determined that we would use a number of previous time steps of motion data (velocity and angular velocity) of the table as inputs. While we had both force and motion data at our disposal from the experiment, only motion data was included in our neural network. Our reasoning for not including force data is that including it caused our neural network solution to have poor convergence. We believe that including a dynamic model, changing the RNN structure, or using a different structure of neural network could help to solve this problem, but we have left this for future work, as we obtained an accurate prediction using our current network with motion data inputs. The basic structure of the neural network is shown in Fig. \ref{fig:structure}.

Our final network consists of 3 LSTM layers each with 100 hidden states. The process of choosing a neural network structure and other
parameters was not exhaustive and it is possible that better structure
and parameters could be obtained. However, the process for choosing neural net parameters and training took over 10 hours. With more experience and more efficient computation, we may have been able to find better structures, but this is also left to future work. Methods other than neural networks may exist, but our purpose in this paper is to show that human intent estimation is possible based on the data collected from our user study, and neural networks allowed us to achieve this goal. 

Additionally, it was shown by Chipalkatty et al. that more complex predictions of future movement can actually decrease performance if they do not agree with what the human is trying to do. This is because humans cannot be completely modelled due to their unpredictability. They found that it was more important that the human understand what the robot will do next, meaning that our controller should be intuitive for a human partner in a human-robot dyad. \cite{Chipalkatty2013} In addition to being intuitive, the prediction should also be accurate and repeatable. The inputs to the neural network, as seen in Fig. \ref{fig:structure}, are 150 past steps of velocity and angular velocity of the table in the x, y, and z directions, $\{x_{t-149}, x_{t-148}...,x_{t-1},x_{t}\}$. The outputs are the predicted velocity and angular velocity of the table in the x, y, and z directions for 1 time step into the future, $\hat{x}_{t+1}$, where $\hat{x}$ indicates a predicted value.

\begin{figure}[t]
  \centering
  \includegraphics[width=.9\linewidth]{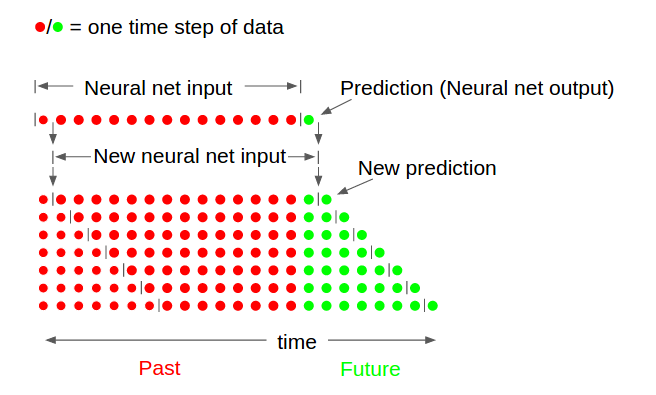}
  \caption{Neural network prediction explanation. Previous time steps are used to obtain one future prediction of states. This state is then appended to previous time steps, the first time step is removed, and the network is run again in order to achieve multiple future predictions.}
  \label{fig:nn_explanation}
\end{figure}

Our neural net formulation also uses what Engel et al. describe as iterated
prediction \cite{Engel2004}. The neural network itself only predicts 1
time step into the future. Then, the prediction, $\hat{x}_{t+1}$, is
appended to the input to give $\{x_{t-149},
x_{t-148}...,x_{t-1},x_{t},\hat{x}_{t+1}\}$. The first step of the input is dropped
to obtain a new input of past motions for the neural net,
$\{x_{t-148}, x_{t-147}...,x_{t},\hat{x}_{t+1}\}$. The new data is
input into the neural net which outputs a prediction 1 step forward,
but 2 total steps into the future, $\hat{x}_{t+2}$. This is then
appended to the input. The process is repeated 50 times to obtain a
prediction of 50 steps, $\{\hat{x}_{t+1},
\hat{x}_{t+2}...,\hat{x}_{t+49},\hat{x}_{t+50}\}$. This process is represented by Fig. \ref{fig:nn_explanation}. Because the outputs
of each prediction step become the inputs for the next, the inputs and
outputs must be the same variables.

\subsection{Training}
We pre-processed the data for the neural net to improve the
results. The velocity and acceleration data were scaled to have 0 mean
and standard deviation of 1 over the entire set of data. This was then
inverted on the output to show the results in their proper units. This
same scaling can be used on new data as long as the mean and standard
deviation are similar to the training data. This is the case in our experiment, as velocity values fall into the average adult human range. The entire set of data consists of 2.5 million time steps for each variable. Data was split into
to training and validation sets. 75\% of the data was assigned to the
training set and the other 25\% to the validation set. 

The neural net has to be trained in a special way in order to make the
iterated prediction $\hat{x}_{t+1}$ stable beyond the first step. This
process, described here, comes from \cite{Engel2004}.  Batches of data were created that randomly pulled in 32 sets of 150 steps of data from the entire training set. Sets of 150 steps were also
created from the validation set. The neural net was trained on new
training batches for a number of iterations, until the cost function was below a threshold we chose. We used the mean squared error (MSE) for the cost function, as shown in Eq. \ref{eqn:definition_Ix}.

\begin{equation} \label{eqn:definition_Ix}
MSE = \sum_{n=1}^{32}(\hat{x}_{n,t+1}-x_{n,t+1})^2 
\end{equation}

Once this threshold was reached, each training batch took in the original data, created a prediction, appended the prediction to the end of the time-series, and removed the first step, following the pattern shown in Fig. \ref{fig:nn_explanation}. This process was repeated until the desired number of future predictions was reached. In other words, the original set of 150 steps was trained to determine a single future step. After this, a new data set was created that used 149 steps of real data and the prediction appended to the end, $\{x_{t-148}, x_{t-147}...,x_{t-1}, x_{t}, \hat{x}_{t+1}\}$. The neural network was then trained on a combined data set that included the original data and the new data set that included the
prediction. Once this training was complete, a new data set was
created with 148 steps of real data and two predictions after it,
$\{x_{t-147}, x_{t-146}...,x_{t-1}, x_{t},\hat{x}_{t+1},
\hat{x}_{t+2}\}$. The same neural net was then trained again, 
continuing until the neural net has predicted the desired number of future time steps. By training the neural network on data that includes
predictions, the stability of the prediction is improved.

The length of the prediction is limited by our computational resources
as we train the neural network, as well as a degradation of prediction after a certain number of steps. Our neural net predicts for 50 steps,
or .25 seconds into the future, because predictions beyond this point did not produce accurate predictions. We speculate this occurs due to
a limit on the predictability of human intent after a certain amount of time. Humans are inherently unpredictable by nature, and we would not expect that an intent estimator could predict an entire trajectory given only a few data points. We also believe that the number of future steps we can predict is dependent on each individual dyad, which we will study more in-depth in the future. Improvements to the neural network architecture may also provide longer prediction times. 
 
The neural net was trained several times randomizing which data was used for training and validation. This ensured that the neural net would generalize to an entirely new dyad, and avoid overfitting. Another benefit of iterated prediction is the inclusion of predicted velocities each training step reduces the amount of overfitting, since new data is essentially being introduced each iteration. By training this neural network, we have created an intent estimator to be used in a human-robot dyad performing co-manipulation of an extended object. The object's motion data can be fed as an input at each time step, and the trained neural network will output a predicted velocity for the object. The robot must then use this estimation to calculate its own trajectory to make the object follow the desired behavior. 

\subsection{Validation}

\begin{figure}[t]
    \captionsetup[subfigure]{justification=centering}
    \centering
    \subfloat[X velocity.]{
        \includegraphics[width=\linewidth]{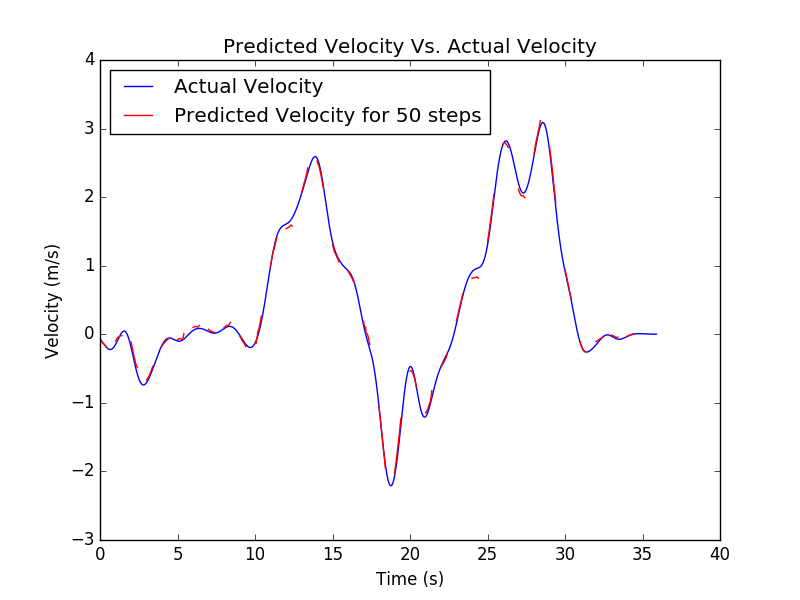}}\\
    \subfloat[Y velocity.]{
        \includegraphics[width=\linewidth]{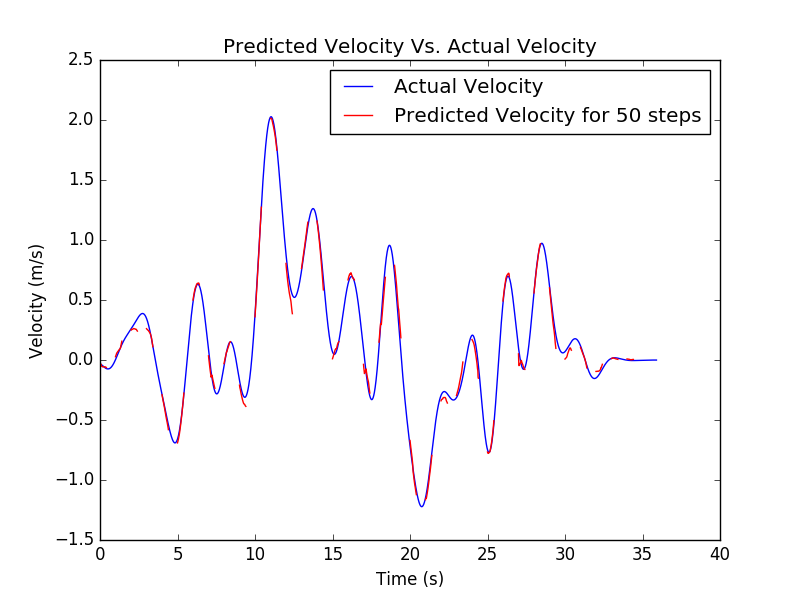}
        }
    \caption{Comparison of velocity prediction in x-direction (a) and y-direction (b) to actual future data while a human dyad moves a table. Each red line is a separate 50 step prediction using the 150 steps before it.}
    \label{fig:prediction}
\end{figure}

Fig. \ref{fig:prediction} shows the neural network predictions of velocity in the x and y directions for a single sequence of the validation set. The actual velocity is shown for the whole task in blue. The predicted velocity is shown in red. Although 50 time steps of prediction were calculated at each time step, the plot only shows the predictions once every second. As seen, the predictions are reasonably accurate for that time scale. Here we only show velocity, but the angular velocity data must also be predicted because each velocity prediction depends on the prediction of angular velocity for the time step before it. Acceleration and angular acceleration data can also be included when training, but is not necessary for accurate predictions.

\subsection{Neural Network Prediction Control}
As discussed previously, and displayed visually in Fig. \ref{fig:control_loop}, predicting human intent is only one portion of the puzzle. We also need to tell the robot what to do with the prediction with a motion controller. The neural network outputs a predicted velocity and angular velocity of the center of mass of the table. Along with the prediction of the velocity of the center of mass, we can know the velocity of other known points on the table, such as the velocity of the leader's edge and the follower's edge, using kinematic relationships. For our motion controller, however, we simply used the predicted velocity of the center of mass, again in the object's reference frame. The robot controller then needs to account for the distance from its center of mass to the table's, which can be done using the transport theorem, shown in Eq. \ref{eq:transport_theorem}. Here, $\vec{v}_r$ is the robot's calculated velocity in its reference frame, with $\vec{p}$ as the distance from the table frame to the robot frame, and $\vec{\omega}$ as the table's angular velocity in the table frame. Also, $\vec{v}_{rel}$ is the table's velocity in its frame. We assumed the table frame and robot frame do not rotate independently, allowing us to rotate the predicted velocities in the table frame to the robot frame.

\begin{equation}\label{eq:transport_theorem}
    \vec{v}_r = \vec{v}_{rel} + (\vec{\omega} \times \vec{p})
\end{equation}
 
\begin{figure}[t]
  \centering
  \includegraphics[width=.85\linewidth]{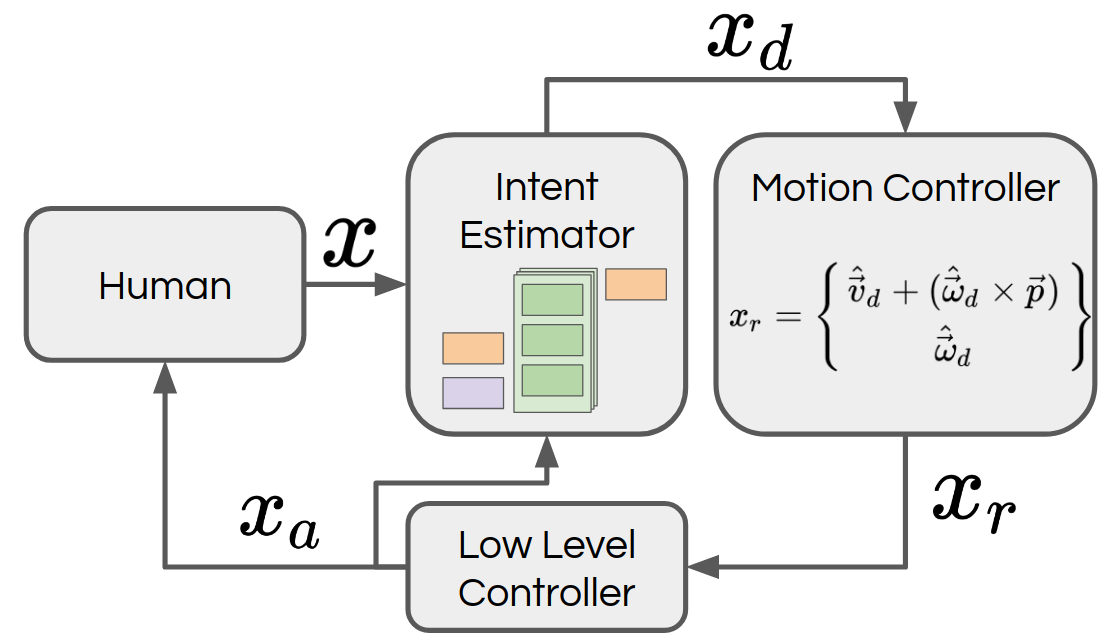}
  \caption{Neural Network Prediction Control loop.}
  \label{fig:NN_loop}
\end{figure}

We now have the tools to complete the control loop shown in Fig. \ref{fig:control_loop}. The intent estimator is replaced with the neural network model. The motion controller is replaced with Eq. \ref{eq:transport_theorem}, and is subsequently fed into the low level control of the robot, which sends voltages down to the wheels to match the desired velocity. The achieved velocity, $x_a$, is then what the human interacts with, completing the loop. Also, $x_a$ is estimated using numerical differentiation and a 2nd order low-pass filter of the pose information coming from the motion capture. This loop is shown in Fig: \ref{fig:NN_loop}. We call this control method Neural Network Prediction Control (NNPC). A notable feature of this method is that the commanded velocity, $x_r$, is a continuous variable on $[-v_{max},v_{max}]$, where $v_{max}$ is determined empirically for each DOF. This means the human user has control of the speed of the interaction, so if the response $x_a$ is not suitable to the human, they can adjust their inputs to move faster or slower.

\begin{figure}[t]
    \captionsetup[subfigure]{justification=centering}
    \centering
    \subfloat[Translation task.]{
    \includegraphics[width=\linewidth]{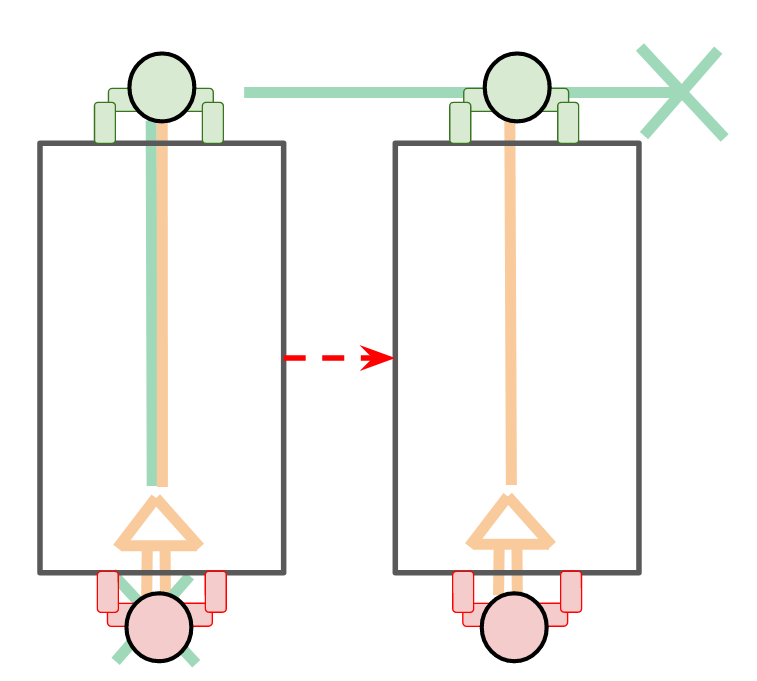}
    }\\
    \subfloat[Rotation task.]{
    \includegraphics[width=\linewidth]{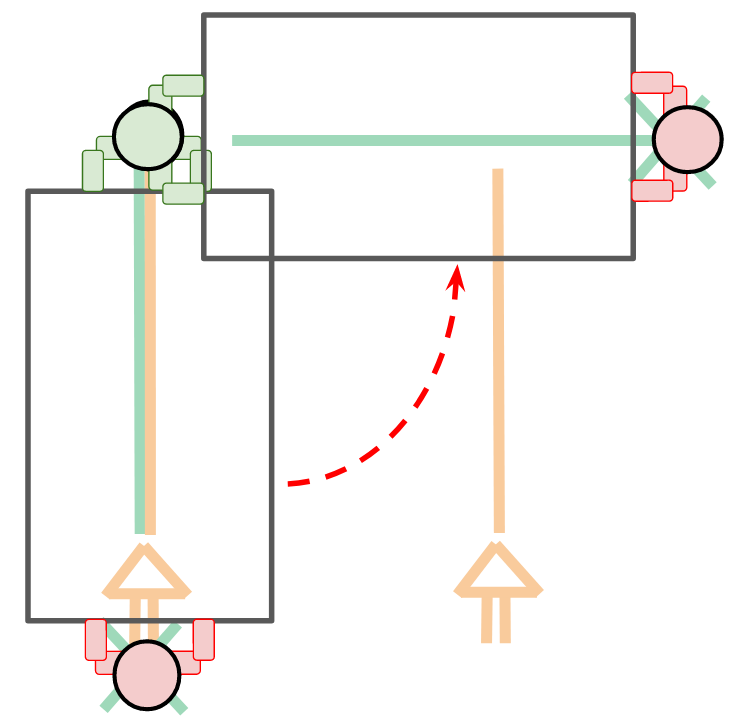}
    }
    \caption{Visual representation of a translation task (a) and a rotation task (b), where the top person represents a robot, and lower person represents a human; tape was used to delineate tasks for user ease with arrows indicating start and stop points for translation task and Xs indicating start and stop points for rotation task}
    \label{fig:trans_and_rot}
\end{figure}

\section{PHRI Co-Manipulation Study}
\label{sec:pilot}
As mentioned in Section, EVIC works only for 3 DOF planar control--anterior and lateral translation and rotation in the plane--so we developed an experiment to compare a planar implementation of NNPC and EVIC.
We believe that since NNPC can provide predictions for all 6 DOF, it can be expanded to control in 6 DOF. However, we have left that for future work as it would also require integration with better robot arm control and is beyond the scope of this paper. This experiment was designed to be as close as possible to the lateral translation and planar rotation tasks from our 6 DOF HHI experiment \cite{Mielke2017}.

\subsection{Experiment Description}
\subsubsection{Tasks}
Because both controllers are planar co-manipulation methods, we used only planar tasks from the previous HHI experiment in \cite{Mielke2017}. As can be seen in Fig. \ref{fig:trans_and_rot}, each participant performed two tasks: translation and rotation. In this diagram, the human is represented by the bottom person. The translation task consisted of the subject moving from one orange arrow to the next, with tape lines extending on the ground to help the user align the board correctly. Rotation tasks were similar, except with the participant moving from one green x to the next. Tasks could be run starting at either arrow or x, and the direction was randomized throughout the trial.

\subsubsection{Equipment}
The position of the board was tracked via Cortex Motion Capture software with a Motion Analysis Kestrel Digital Realtime System. A total of 8 Kestrel cameras were used to track 8 infrared markers placed on the board.  Using a static global frame established by the motion capture system, the position and orientation of the board could be tracked over time, and we transformed the data into the robot's frame for use in the neural network. The motion capture data was collected at a rate of 200 Hz. In order to run NNPC, we need a method of telling the controller the object velocity. We used a 2nd order low-pass filter and numerical differentiation on position and orientation data to get the object velocity. Additionally, participants wore sleeves with infrared markers to track the position of their arms during the trial. This data was not used during analysis, but was collected to match similar data collected during the experiment in \cite{Mielke2017}.

The object the teams moved was a 59x122x2 cm wooden board -- meant to simulate an object (like a table) that is difficult for one person to maneuver, which weighs 10.3 kg. Attached to the robot end of the board were a pair of ABS 3D-printed handles, to which two ATI Mini45 force/torque sensors were fastened. The sensors transmitted data via ATI NET F/T Net Boxes, which passed data over Ethernet to the computer at a rate of 100 Hz. The sensor is attached to wrist adapters on the other side, which fasten to Baxter's wrists.

The test arena was a volume measuring 490x510x250 cm. The arena was also equipped with a video capturing device. The device we used was a Microsoft Kinect 2, which allowed us to capture 3D point cloud data, as well as color video of each trial. Although we did not use the point cloud data for analysis in this paper, the data may be useful in future work.

\subsubsection{Subjects and Procedure}
Subjects for this study were male and female students from Brigham Young University in Provo, UT. There were a total of 16 students--4 female and 12 male--ranging from 18-23 years of age, with an average age of 20. Students were from a variety of majors, with STEM majors making up a majority. Participants were asked to rate their familiarity with robots on a scale from 1-5, with 5 being the most familiar, and the average rating was 2. IRB approval was obtained for this experimental study.

Participants entered the Robotics and Dynamics Lab, and provided written informed consent in accordance with IRB. They were then briefed on the purpose of the research and given an introduction to what data would be collected, and what would be expected of them. Sleeves were then placed on the participants arms in order to track their arm motion during the trial. Subjects were then given basic operating instructions for both EVIC and NNPC controllers. This instruction included how to translate in the anterior and lateral directions, and how to rotate the board for each controller. A controller was randomly selected, and each participant practiced with that controller until they were able to complete a competency task, they moved on to the other controller, and repeated the competency training. The competency task consisted of aligning the board with the tape lines on the ground, starting from a translated and rotated position. The practice assured us that each participant would have at least enough familiarity to complete the translation and rotation tasks.

Once competency training was completed, a controller was selected at random to be the first controller for data collection. The randomization of controllers was counterbalanced. Participants knew the controllers only as A (NNPC) and B (EVIC). They were not given any specific details about the formulation of the controllers, other than the basic operating instructions in the competency task. The subjects then ran a series of translation and rotation tasks with the selected controller. Tasks were randomized (counterbalanced) in order between translation and rotation. Once a type of task, either rotation or translation, was selected, the participant ran that task type one direction, and then ran the same task type, but in the other direction. The direction order was not randomized, and the first direction is indicated in Fig. \ref{fig:trans_and_rot}. Due to the nature of the controller, the robot was not able to lift the table from the ground, so the table was laid on a rest stand between trials. A single trial consisted of the subject lifting the table off of the rest, then a researcher would remove the rest from below the table. Once the rest was completely out of the way, the subject then performed the specified task. Participants indicated they were finished by verbally communicating completion. Once they indicated they had completed the task, a researcher would replace the rest underneath the table, and the participant would lower the table back onto the rest. Each task was repeated 6 times, 3 one direction and 3 the other direction, for each controller. Once trials were completed for one controller, the participants were given a survey, and asked to rate the controller on certain qualitative characteristics. Once completed, they moved on to the other controller.

A video showing a representation of tasks can be seen online at \url{https://youtu.be/4b-wxn9_gFQ}. This video was taken after the participant had completed all trials, so we could use a higher definition camera to record video, and what is seen is a good representation of the skill level of the human-robot dyad post-experiment.
 
\section{Results and Discussion}
\label{sec:results-and-discussion}

\subsection{Metrics}
There are a number of metrics that can be used to quantify performance of the controllers. A good summary of these metrics is found in \cite{Ivaldi2012}. Among these metrics are a few that are especially applicable to the tasks and control methods we have developed. These are: minimum jerk, minimum torque change, and completion time. While none of these metrics alone can completely store all the information of each controller, together they give us a fairly good indication of how each controller was performing in relation to HHI data from \cite{Mielke2017}. 

Minimum jerk error (MJE), or deviation from a minimum-jerk trajectory, is a measure of how close the actual trajectory was to a minimum-jerk trajectory in meters, and is calculated using Eq. \ref{eq:mje}. This measure accounts for human tendency to match these trajectories. Completion time is simply the time from the start of the task to the end of the task. The start of the task is calculated by determining when the object has first moved beyond 5\% of the distance between the starting and ending $y$ or $\theta_z$ positions, for translation and rotation respectively. The end of the task is calculated by determining when the object settles into 95\% of the distance between starting and ending $y$ or $\theta_z$ positions. A buffer of 0.5 s is added to the total time to account for the missed motion. This measure accounts for how quickly a dyad performed the task. Although quick task completion is not always a direct objective of dyads, this measure can help compare the capabilities of two dyads. Minimum-torque measure (MTM) is a measure of how much the time-derivative of torque changes over the course of the task. In instances where the follower predicted incorrectly, there was an unforeseen obstacle, or some other disturbance, MTM can account for human tendency to reduce the amount of force or torque required to move. It is calculated using Eq. \ref{eq:mtm}.

\begin{equation}\label{eq:mje}
    MJE = \sum_{t=0}^{T} x_{mj,t} - x_{a,t}
\end{equation}

\begin{equation}\label{eq:mtm}
MTM = \sum_{t=0}^{T-1}\dot{\tau}_{t}^2 + \dot{\tau}_{t+1}^2
\end{equation}

\subsection{Quantitative Results}
While each task type was performed 6 times for each controller, we only consider the data from the last 2 trials performed for each task type. Our reasoning behind this is that each participant learned throughout the experiment, so we want to take the best representation of the control method. This is acceptable to us, because if a human-robot team were to be deployed in real life, the human would undoubtedly be trained on working with the robot. If we consider the practice time and time taken to perform the first 4 trials of each type, the total training time for our participants was approximately 30 minutes. 

The results from the experiment based on the metrics in the previous section can be seen in Table \ref{table:metrics}. This table compares EVIC and NNPC to each other, as well as to the lower (blind HHI) and upper (sighted HHI) bounds of human performance. As can be seen, NNPC performed best in most of the metrics. It was able to get closer to blind HHI performance in completion time, and outperformed EVIC, as well as both blind and sighted HHI performance in both MJE and MTM, where lower numbers indicate more efficient performance in that task. EVIC, while not quite as good, still outperformed blind and sighted HHI in most of the metrics, except for completion time. It is notable that the blind HHI performance captured here is for a human-human leader-follower dyad, where the follower was blindfolded, and communication was limited to haptic communication only, whereas sighted HHI allowed for communication in any form desired by the dyad.

\bgroup
\def\arraystretch{1.25}
\begin{table*}[t]
	\fontsize{11}{9}
	\caption{Metrics of EVIC and NNPC for rotation and translation tasks, also compared with blindfolded HHI and sighted HHI data from \cite{Mielke2017}}
	\label{table:metrics}
\begin{center}
	\begin{tabular}{| c | c | c | c | c |}
	\hline
	Metric and Task Type & Blind HHI & EVIC & NNPC & Sighted HHI\\
	\hline
	Completion Time (s)--Rotation  & 7.08 & 8.25 & 8.26 & 6.58\\
	\hline
	Completion Time (s)--Translation & 7.18 & 7.91 & 7.75 & 4.93\\
	\hline
	MJE (rads)--Rotation  & 392.71 & 96.44 & 87.38 & 344.70\\
	\hline
	MJE (m)--Translation & 149.91 & 50.24 & 48.51 & 98.92\\
	\hline
	MTM (N\textsuperscript{2}$\cdot$m\textsuperscript{2}/s\textsuperscript{2})--Rotation  & 488454.38 & 65602.60 & 12770.75  & 341253.43\\
	\hline
	MTM (N\textsuperscript{2}$\cdot$m\textsuperscript{2}/s\textsuperscript{2})--Translation & 387937.56 & 48191.90 & 15220.89 & 151758.83\\
	\hline
	\end{tabular}
\end{center}
\end{table*}
\egroup

We also determined the statistical significance of these quantitative results. The mean values of the metrics for both the EVIC and NNPC controllers were compared with each other, and also with the blind and sighted human-human dyads, based on the 3 metrics used in this section. We ran an unpaired t-test to determine p values, and also determined Cohen's d to calculate an effect size. From the p values, we were able to see which groups were statistically likely to have the same mean, and therefore see whether there was a statistical difference between the groups, based on a standard $p<0.05$ criteria. With the effect size, we were able to determine the strength of each comparison. Effect sizes were calculated, and then categorized into very small, small, medium, large, very large, or huge categories, based on Sawilowsky's work \cite{Sawilowsky2009}. These statistics are summarized in Table \ref{table:stat_sig_quant}.

\bgroup
\def\arraystretch{1.25}
\begin{table*}[t]
	\fontsize{11}{9}
	\caption{Statistical significance of quantitative metrics.}
	\label{table:stat_sig_quant}
\centering
	\begin{tabular}{| c | c | c | c | c | c | c |}
	\hline
	& \multicolumn{3}{|c|}{$p$} & \multicolumn{3}{|c|}{Cohen's d Effect Size}\\
	\hline
    Comparison Groups & Comp. Time & MJE & MTM & Comp. Time & MJE & MTM \\
	\hline
	EVIC vs NNPC Trans. & 0.73 & 0.89 & 0.017 & Small & Medium & Large \\
	\hline
    EVIC vs NNPC Rot. & 0.98 & 0.70 & 0.14 & Very Small & Small & Medium \\
	\hline
	EVIC vs Blind Trans. & 0.07 & 0.00 & 0.00 & Medium & Huge & Huge \\
	\hline
	EVIC vs Blind Rot. & 0.06 & 0.00 & 0.00 & Medium & Very Large & Very Large \\
	\hline
	NNPC vs Blind Trans. & 0.16 & 0.00 & 0.00 & Medium & Huge & Huge \\
	\hline
	NNPC vs Blind Rot. & 0.05 & 0.00 & 0.00 & Medium & Very Large & Very Large \\
	\hline
	EVIC vs Sighted Trans. & 0.00 & 0.00 & 0.00 & Huge & Large & Very Large \\
	\hline
	EVIC vs Sighted Rot. & 0.01 & 0.00 & 0.00 & Large & Large & Large \\
	\hline
	NNPC vs Sighted Trans. & 0.00 & 0.00 & 0.00 & Huge & Large & Huge \\
	\hline
	NNPC vs Sighted Rot. & 0.01 & 0.00 & 0.00 & Large & Large & Very Large \\
	\hline
	\end{tabular}
\end{table*}
\egroup

There are a few results which are important to recognize from this analysis. First, EVIC and NNPC are not statistically different in terms of completion time or MJE, but do seem to differ in MTM, which additionally has a fairly large effect size. Second, both EVIC and NNPC are not statistically different from the blind human-human dyads in terms of completion time. Last, EVIC and NNPC are statistically different from both blind and sighted human-human dyads in terms of minimum-jerk error and MTM, and these comparisons are all categorized as large or higher. A more in-depth discussion of these results is found in Section \ref{sec:discussion}, but as a short summary, the statistics show that these controllers have approached a level comparable to blind human-human dyads with respect to the completion time metric, but are still quite distinguishable in terms of MJE and MTM metrics.

\begin{figure}[t]
  \centering
  \includegraphics[width=1.0\linewidth]{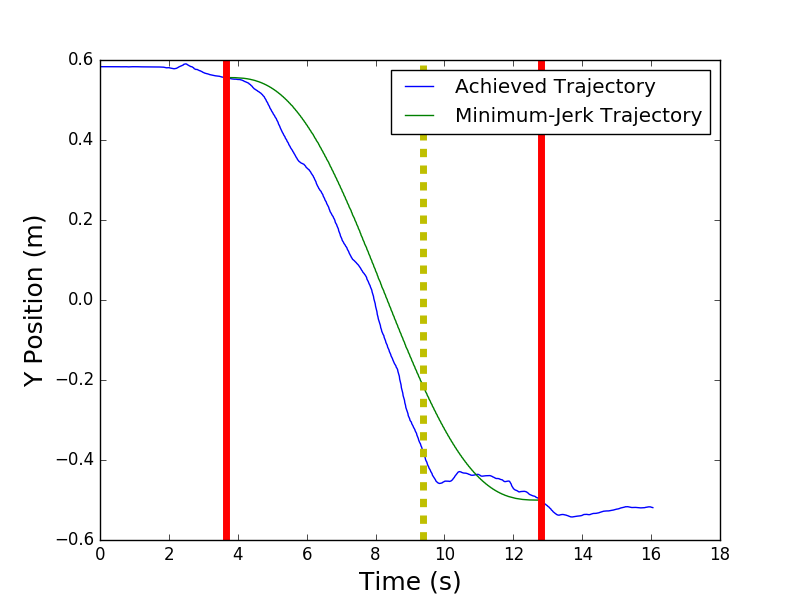}
  \caption{Undershooting behavior of a human robot dyad for a translation task, where bold, vertical lines indicate start and stop points, and dashed vertical line indicates 90\% completion point--movement after this point is considered a fine motor adjustment}
  \label{fig:fine}
\end{figure}

Another noteworthy observation is that both EVIC and NNPC, while capable, have difficulties with fine-motor adjustments. Throughout the trials, participants occasionally overshot or undershot their desired position, and had to make fine motor adjustments to achieve the desired position. An example of undershooting is shown in Fig. \ref{fig:fine}. The dyad is able to complete 90\% of the task, represented by the dashed vertical line, in just under 6 seconds, but spends approximately 3 seconds trying to complete the remaining 10\%, which amounts to about 10 cm of movement, with more fine adjustments. 

On average, the remaining time at 90\% completion was 2.40s for EVIC and 2.55s for NNPC. We also compared the achieved trajectories with an ideal minimum-jerk trajectory. The time to 90\% task completion on a minimum-jerk trajectory was, on average, 2.20s for EVIC and 2.12s for NNPC. From this data, it would appear that EVIC is slightly better at fine-motor adjustments than NNPC, since EVIC had a smaller discrepancy between achieved and minimum-jerk 90\% completion time. 

To determine if a few under-performing dyads skewed the average, we also took the median 90\% completion time. For achieved and minimum-jerk trajectories, respectively, with EVIC, this gave values of 1.98s and 2.11s. Similarly for NNPC, it gave values of 1.97s and 2.02s. From these results, we conclude that this data is positive skewed, and only a small number of dyads had trouble with fine-motor adjustments, causing the higher mean values. Therefore, we can conclude the lack of fine motor skills in the controllers did not significantly hamper their ability to complete the tasks, but should be addressed in future work to help improve the performance of those dyads who struggled with undershooting or overshooting.

\subsection{Qualitative Results}
In addition to the metric performance, we also wanted to know what each person thought of the controllers subjectively. As mentioned in Section \ref{sec:pilot}, we gave each participant a survey after they had performed all the tasks with one controller, and then the same survey after they finished the other controller. The questions we asked dealt with how they thought their partner, a robot in this case, performed. They were asked to rate the categories from 1--Strongly Disagree--to 5--Strongly Agree. The questions were: 

\begin{itemize}
    \item My partner was helpful
    \item My partner moved quick enough
    \item My partner moved too slow
    \item There was confusion between me and my partner
    \item I trusted my partner to perform the task correctly
    \item I felt safe completing the task
    \item I trusted my partner to move at correct speeds
    \item I trusted my partner to move in the correct direction
    \item My partner did not push or pull too hard
    \item My partner moved predictably
    \item My partner helped me to do task better than I could alone
    \item My partner shared the task equally
\end{itemize}

The average for each controller rating is given in the first 2 columns of Table \ref{table:survey}. The controller that performed better in each category is in bold. For some categories, like \textit{Too Slow}, a lower number is desired, whereas for others, like \textit{Safe}, a higher number is desired. The same survey, except for the \textit{Correct Direction} question, was given after the HHI study, and the results are shown in the third column of Table \ref{table:survey}. We have included only the responses of the human designated as the leader from the human-human dyads. 

\bgroup
\def\arraystretch{1.25}
\begin{table}[t]
	\fontsize{11}{9}
	\caption{Ratings of survey questions, with 5 as strongly agree and 1 as strongly disagree. Bold numbers indicate preference between EVIC and NNPC for the specified category}
	\label{table:survey}
\begin{center}
	\begin{tabular}{| c | c | c |}
	\hline
	& EVIC & NNPC \\
	\hline
	Helpful & \textbf{\small{3.88}} & \textbf{\small{3.88}} \\
	\hline
	Fast Enough & 3.38 & \textbf{\small{3.63}} \\
	\hline
	Too Slow & 3.31 & \textbf{\small{2.94}} \\
	\hline
	Confusing & \textbf{\small{3.38}} & 3.5 \\
	\hline
	Correct Task & \textbf{\small{3.94}} & 3.75 \\
	\hline
	Safe & \textbf{\small{4.5}} & 4.44 \\
	\hline
	Correct Speed & 3.44 & \textbf{\small{3.56}} \\
	\hline
	Correct Direction & 3.5 & \textbf{\small{3.56}} \\
	\hline
	Good Force Amount & \textbf{\small{3.44}} & 2.81 \\
	\hline
	Predictable & \textbf{\small{3.63}} & 3.5 \\
	\hline
	Better than Alone & 3.5 & \textbf{\small{3.56}} \\
	\hline
	Equal Share & \textbf{\small{3.75}} & 3.5 \\
	\hline
	\end{tabular}
\end{center}
\end{table}
\egroup

\bgroup
\def\arraystretch{1.25}
\begin{table}[t]
	\fontsize{11}{9}
	\caption{Statistical significance of qualitative metrics.}
	\label{table:survey_stat_sig}
\begin{center}
	\begin{tabular}{| c | c | c |}
    \hline
	& \multicolumn{2}{|c|}{EVIC vs NNPC}\\
	\hline
	Survey Question & $p$ & Cohen's d Effect Size \\
	\hline
	Helpful & 0.5 & Very Small \\
	\hline
	Fast Enough & 0.19 & Medium \\
	\hline
	Too Slow & 0.15 & Medium \\
	\hline
	Confusing & 0.34 & Small \\
	\hline
	Correct Task & 0.28 & Medium \\
	\hline
	Safe & 0.36 & Small \\
	\hline
	Correct Speed & 0.37 & Small \\
	\hline
	Correct Direction & 0.41 & Small \\
	\hline
	Good Force Amount & 0.04 & Medium \\
	\hline
	Predictable & 0.35 & Small \\
	\hline
	Better than Alone & 0.38 & Small \\
	\hline
	Equal Share & 0.15 & Medium \\
	\hline
	\end{tabular}
\end{center}
\end{table}
\egroup

Similar to our analysis of the quantitative results, we determined the statistical significance of the survey results, in order to determine if there was a statistical difference between the responses about the EVIC and NNPC controllers. For each question, we ran an unpaired t-test to calculate a p value and determined Cohen's d to find the effect size. These statistics are found in Table \ref{table:survey_stat_sig}. What we found is that only the \emph{Good Force Amount} question obtained a p value of less than 0.05, indicating it was the lone statistically significant answer. However, this question, as well as a number of others, had a medium effect size.

\subsection{Discussion}\label{sec:discussion}
The first thing we noticed from Table \ref{table:survey} is that people still clearly prefer working with a human partner over a robot partner. One reason for this may be that humans do not trust robots entirely, as is evidenced by the 5th, 7th, and 8th questions in the survey, which all ask about trust in the partner. Perhaps the same pHRI experiment, done instead with a blindfold and earmuffs on the human would have returned more favorable ratings for the robot controllers. This is something that needs to be explored further, as our experiments do not definitely prove that humans do not trust robots.

As was mentioned above, NNPC was the more capable controller in terms of the performance metrics. It also was preferred in the survey questions that dealt with these metrics. NNPC was superior in rating for the \textit{Fast Enough, Too Slow, Correct Speed, Correct Direction,} and \textit{Better than Alone} metrics. The superior rating in these metrics corroborates what we saw quantitatively, indicating that NNPC was a more efficient controller, at least according to the metrics we have seen. 

However, participants also indicated they preferred using EVIC. They thought it was more helpful, applied more appropriate levels of force, and saw it as the more predictable controller. While NNPC outperformed EVIC in the TMT metric, this may not be a good thing. From our observations in this paper, haptic communication is a large part of how humans perform co-manipulation tasks. NNPC users experienced less force overall, based on the TMT metric, but the survey indicated that EVIC applied more appropriate forces. Additionally, EVIC and NNPC were only statistically different in terms of the MTM metric and the \emph{Good Force Amount} question. From these results, we can conclude that NNPC is not applying appropriate forces, and is therefore considered more difficult and less intuitive to use by the participants. In fact, our results here agree with what was said by Chipalkatty et al. \cite{Chipalkatty2013}, who indicated that training a controller to be the most efficient or best performing controller may cause it to be a less preferable controller to humans. Because EVIC and NNPC were statistically different from both blind and sighted human-human dyads in terms of MJE and MTM, we can say that minimizing forces and torques, and also deviation from a minimum-jerk trajectory, may not be a satisfactory goal of co-manipulation controllers. So while NNPC is the better performing controller, EVIC might be a more intuitive and appropriate controller for real-world applications, since it applies more appropriate forces. 

Looking at completion time in Table \ref{table:stat_sig_quant}, we can see that both EVIC and NNPC were not statistically distinguishable from the blind human-human dyads. This means that according to this metric, both controllers perform up to the standard of the blind dyads. While this is an encouraging result, We know that there is some missing information in our model. Although similar in completion time, our controllers performed much different than both types of human-human teams in the TMT and MJE metrics.  Perhaps there is a complex cost function that humans are using to determine how much weight to put on adhering to minimum-jerk trajectories, minimizing torque change, and moving in a timely manner. These considerations should be made in future controller development.
 
We have also seen (see Fig. \ref{fig:fine}) that these controllers are not well suited for fine-motor adjustments. We believe that because our robot arms were not helping to reach position, but were simply holding an equilibrium position, the dyad was not able to account for small errors in the position of the board. An algorithm that is able to coordinate arm and base motion would be more suited to fix these small errors, and could possibly bring our pHRI controllers for co-manipulation closer to human-human sighted performance.

\section{Conclusion}
\label{sec:conclusion}
Despite the progress we have made related to the problem of human-robot co-manipulation, there remains signficant future work. Future work involves incorporating force into NNPC, either through a more capable neural network, through combining NNPC and EVIC, or through a different approach, such as ARMAX. We also want to test NNPC in higher DOF, to see how capable it is in tasks that require more DOF than the human-robot experiment designed here. Many real world tasks will involve at least lifting and object up off the ground into the plane, before switching to a relatively planar motion. We also want to see how NNPC performs in other tasks described in Section \ref{human_study}, to see if it is capable of performing those tasks. This will involve creating a more sophisticated arm controller for Baxter, which may involve studying the arm movements of human-human dyads from our exisiting data sets to characterize their movement.

In this paper, we have discussed the problems and limitations of many current co-manipulation pHRI controllers, especially as their limitations relate to co-manipulation of extended objects. We discussed the advantages of creating control methods based on human-human dyad behavior to increase the ability of human-robot dyads to adapt to less-defined situations. We also described our experiment gathering the force and motion data for several simple and complex tasks involving human dyads. The main results from analyzing this data include that interaction forces play an important role in communicating intent between dyads in co-manipulation and that they are likely not minimized as previously supposed. Planar movements display characteristics of minimum-jerk movements, and deviation from MJ trajectories could be used as a metric for extended object co-manipulation. 

We also discussed our implementation of Extended Variable Impedance Control, a novel method for planar 3 DOF co-manipulation of extended objects, and its advantages over standard Variable Impedance Control, as well as Bi-Manual Variable Impedance Control, an extension of a controller from related work. 

Finally, we have also shown that human intent can be estimated accurately from previous motion of the object that is being co-manipulated. We have shown that a RNN with velocity inputs is capable of capturing human intent in the form of velocity estimation. We have implemented a neural-network based controller on a Baxter Research Robot mounted onto a AMP-I Mobile Base, running the neural network control to command the wheels, and using a simple impedance controller to maintain arm position. We described an experiment to compare our neural network controller, NNPC, to a previous method, EVIC, and also to blind HHI, or human-human dyads with a blindfolded follower and only haptic communication, performance, as well as sighted HHI performance. We found that NNPC outperformed EVIC in all metrics considered, and that both NNPC and EVIC were comparable to blind human-human dyads in completion time. We also found that although NNPC was the superior controller based on performance, participants preferred EVIC, claiming they felt it was safer, less confusing, and more predictable. We believe that NNPC sacrifices some intuition for performance, and the added performance capabilities are unfamiliar to human users, and they feel less comfortable than with the force-based EVIC.

\section*{Compliance with Ethical Standards}
{\bf Funding:} This work was supported by the Army RCTA program. All results and conclusions are the responsibility of the
authors and do not necessarily reflect the opinions of the funding source.

\noindent
{\bf Ethical approval:} All procedures performed in studies involving human participants were in accordance with the ethical standards of the institutional and/or national research committee and with the 1964 Helsinki declaration and its later amendments or comparable ethical standards.

\noindent
{\bf Informed consent:} Informed consent was obtained from all individual participants included in the study.

\section*{Acknowledgements}
We acknowledge and are grateful for helpful conversations with Michael Goodrich, Steven
Charles, and Ryan Farrell (all from Brigham Young University) about our human-human experiments.

This work was funded by the United States Army Research Laboratory through contract W911NF-14-1-0633.


\bibliography{hr_comanip}
\bibliographystyle{IEEEtran}

\end{document}